%% file: main.tex
\documentclass{article} % For LaTeX2e
\usepackage{iclr2023_conference,times}

\input{math_commands.tex}

\usepackage{calc}
\usepackage{hyperref}
\usepackage{url}
\usepackage{times}
\usepackage{amsmath}
\usepackage{enumitem}
\usepackage{latexsym}
\usepackage{xspace}
\usepackage{graphicx}
\usepackage{tabularx}
\usepackage{booktabs}
\usepackage{subcaption}
\usepackage{tikz}
\usepackage{placeins}
\usepackage{multirow}
\usepackage[T1]{fontenc}
\usepackage[utf8]{inputenc}
\usepackage{xcolor}
\usepackage{lipsum}
\usepackage{tikz}
\usepackage{xparse}
\usepackage{amssymb}% http://ctan.org/pkg/amssymb
\usepackage{pifont}% http://ctan.org/pkg/pifont
\usepackage{cancel}
\usepackage{scalerel}

\input{macros}

\iclrfinalcopy

\input{title}

\author{Tao Li \; Gang Li \; Jingjie Zheng \; Purple Wang \; Yang Li \\
Google Research, Mountain View, U.S.A.\\
{\tt \footnotesize \{tlinlp,leebird,jingjiezheng,purplewang,liyang\}@google.com}
}

\begin{document}

\maketitle

\input{abstract}

\input{intro}

\input{background}

\input{task}

\input{data}

\input{modeling}

\input{experiments}

\input{conclusion}

%\bibliography{iclr2023_conference}
\bibliography{anthology,custom}
\bibliographystyle{iclr2023_conference}

\appendix
\input{appendix}

\end{document}

%% file: math_commands.tex
\usepackage{amsmath,amsfonts,bm}

\def\eqref#1{equation~\ref{#1}}
\def\1{\bm{1}}

\DeclareMathAlphabet{\mathsfit}{\encodingdefault}{\sfdefault}{m}{sl}
\SetMathAlphabet{\mathsfit}{bold}{\encodingdefault}{\sfdefault}{bx}{n}

%% file: macros.tex
\newcommand{\myname}{\textsc{Mug}\xspace}
\newcommand{\foneat}[1]{\text{F1}\textsubscript{#1}}

\newcommand{\tb}[1]{\textbf{#1}}

\newcommand{\blue}[1]{{\color{blue} \textit{#1}}}
\newcommand{\red}[1]{{\color{red} \textit{#1}}}
\newcommand{\gold}[1]{{\color{brown} \textit{#1}}}

\newcommand{\cmark}{\ding{51}}
\newcommand{\xmark}{\ding{55}}

\newlength\myheight
\newlength\mydepth
\settototalheight\myheight{Xygp}
\newcommand*\inlinegraphics[1]{%
  \settototalheight\myheight{Xygp}%
  \settodepth\mydepth{Xygp}%
  \raisebox{-\mydepth}{\includegraphics[height=\myheight]{#1}}%
}

%% file: title.tex
\title{\myname: Interactive Multimodal Grounding on User Interfaces}

%% file: abstract.tex
\begin{abstract}
%We present \myname, a novel task of Multi-tUrn Grounding where a human user and the agent (model) work interactively to locate the target element on a user interface (UI) screen. %User interface understanding is often formulated as a grounding task where an agent associates vision component to a user's natural language instruction.
%Prior UI grounding works often rely on a single command from the user. Yet, in a realistic situation, user commands might be ambiguous and the target on a graphical UI can be inherently difficult to articulate verbally. \myname allows multiple rounds of interactions where the agent selects an element in response to a user command and a user gives further commands to correct or refine the selection of the agent. Such interactions allow the model to leverage a history of commands and selections and are critical for improving grounding accuracy thus befit real-world use cases. To investigate the problem,
%we create a new dataset that consists of $77,820$ examples of user-agent interactions on mobile interfaces. To establish a benchmark, we experiment with a set of models for the task and propose several new metrics for automatically evaluating model performances. Our experiments show that incorporating multi-turn interaction substantially improved model grounding performance, and our proposed metrics highly correlated with human-based evaluation.

We present \myname, a novel interactive task for multimodal grounding where a user and an agent work collaboratively on an interface screen.
Prior works modeled multimodal UI grounding in one round: the user gives a command and the agent responds to the command.
Yet, in a realistic scenario, a user command can be ambiguous when the target action is inherently difficult to articulate in natural language.
\myname allows multiple rounds of interactions such that upon seeing the agent responses, the user can give further commands for the agent to refine or even \emph{correct} its actions. Such interaction is critical for improving grounding performances in real-world use cases. To investigate the problem, we create a new dataset that consists of $77,820$ sequences of human user-agent interaction on mobile interfaces in which $20\%$ involves multiple rounds of interactions.
To establish our benchmark, we experiment with a range of modeling variants and evaluation strategies, including both offline and online evaluation---the online strategy consists of both human evaluation and automatic with simulators.
%As a seemingly easy single-screen grounding task, 
Our experiments show that allowing iterative interaction significantly improves the absolute task completion by 18\% over the entire test set and 31\% over the challenging subset.
Our results lay the foundation for further investigation of the problem.

%interactively correcting agent error is a challenging task. While incorporating interaction substantially improves grounding task completion from $69\%$ to $75\%$ at best, when they are instructed to correct themselves, grounding models are prone ($\textt{\geq}37\%$) to repeat the same errors.

\end{abstract}

%% file: intro.tex
\section{Introduction} \label{sec:intro}

% A modified version below this par
%Natural language understanding on graphical user interfaces (GUIs) is crucial for realizing conversational interaction and addressing interaction scenarios such as accessibility where language is a important modality. Specifically, interpreting user commands into executable actions has drawn increasing interests of the research community as it manifests rich research problems including multimodal modeling and natural language grounding. Although substantial progress has been made in the topic, prior work has mostly considered this problem in a single-pass fashion where the model predicts actions based on a given command or instruction. However, in a realistic scenario, a user command can be ambiguous or inaccurate, e.g., when the target action is inherently difficult to articulate or the user command is casual or high level. As a result, it is important to allow interactions in the grounding process where the user has the opportunity to clarify their intention and the model (agent) to revise and correct its action.

Natural language understanding on graphical user interfaces (GUIs) is crucial for realizing human-computer interaction and assisting scenarios that have accessibility difficulties~\citep{Sarsenbayeva2018-ux}.
Specifically, interpreting user commands into executable actions has drawn increasing interests
%of the research community
as it manifests rich research problems including multimodal modeling and natural language grounding~\citep[e.g.,][]{sugilite-toby, gur2018learning, He2020-nq, Li2020-gl, Li2021-re}.
Prior works often consider UI grounding in a single-pass fashion where the model predicts actions with a given instruction
without looking backward to refine prediction.
%, and if error happens, the process starts over again.
However, in a realistic scenario, user instructions can be ambiguous or inaccurate especially when the target action is difficult or inconvenient to articulate.
Reasoning in such cases is inherently iterative.
Therefore, it is important and beneficial to incorporate interaction for resilient grounding~\citep{suhr-etal-2019-executing, chandu-etal-2021-grounding}.

\begin{figure*}[t!]
    \centering
\scalebox{0.8}{
  \begin{subfigure}[b]{0.55\textwidth}
    \includegraphics[height=0.3\textheight]{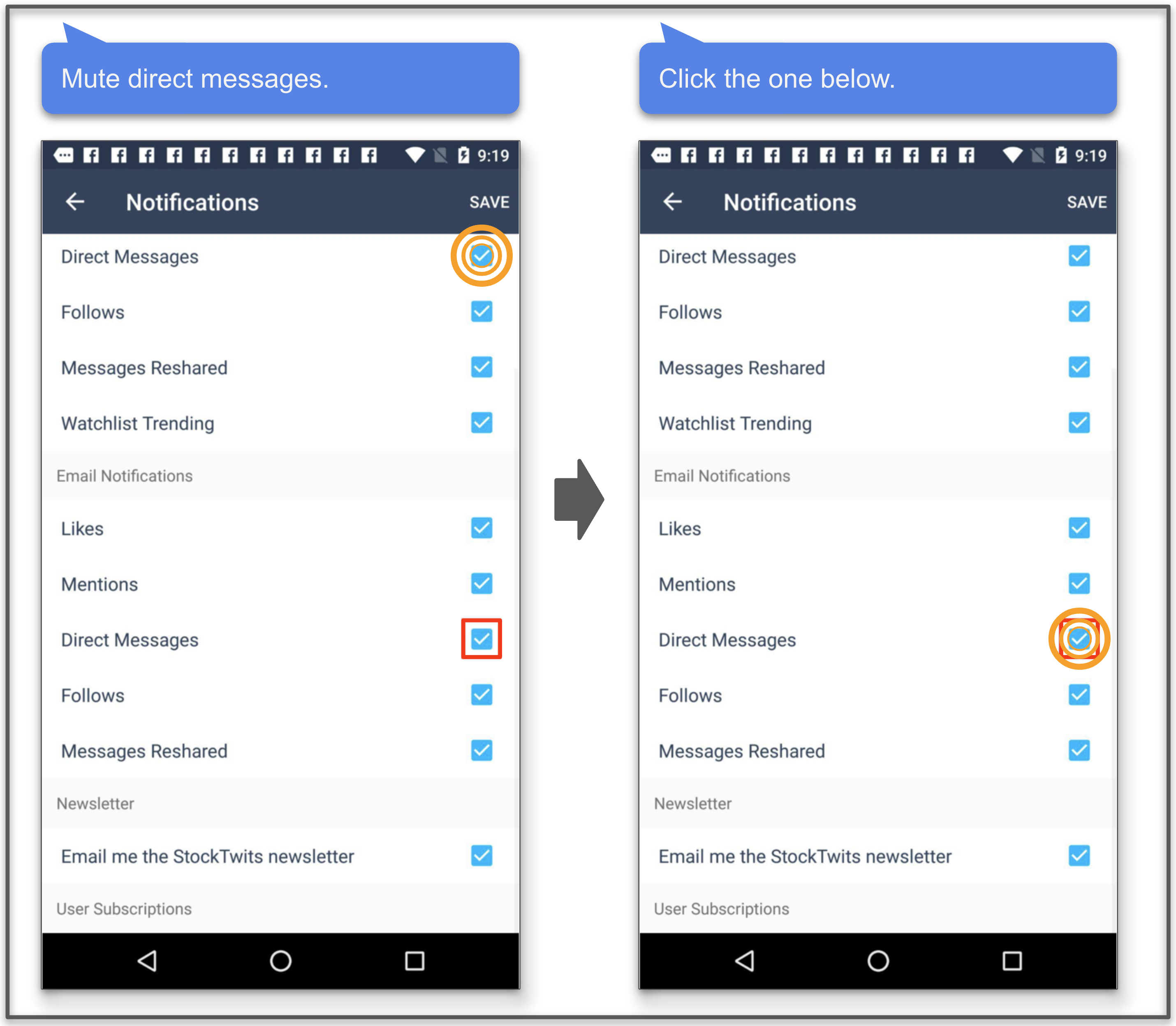}
    \caption{\small{Example one.}}
    \label{fig:intro_ex1}
  \end{subfigure}%
  \hspace{2em}
  %\hspace*{\fill}   % maximize separation between the subfigures
  \begin{subfigure}[b]{0.55\textwidth}
    \includegraphics[height=0.3\textheight]{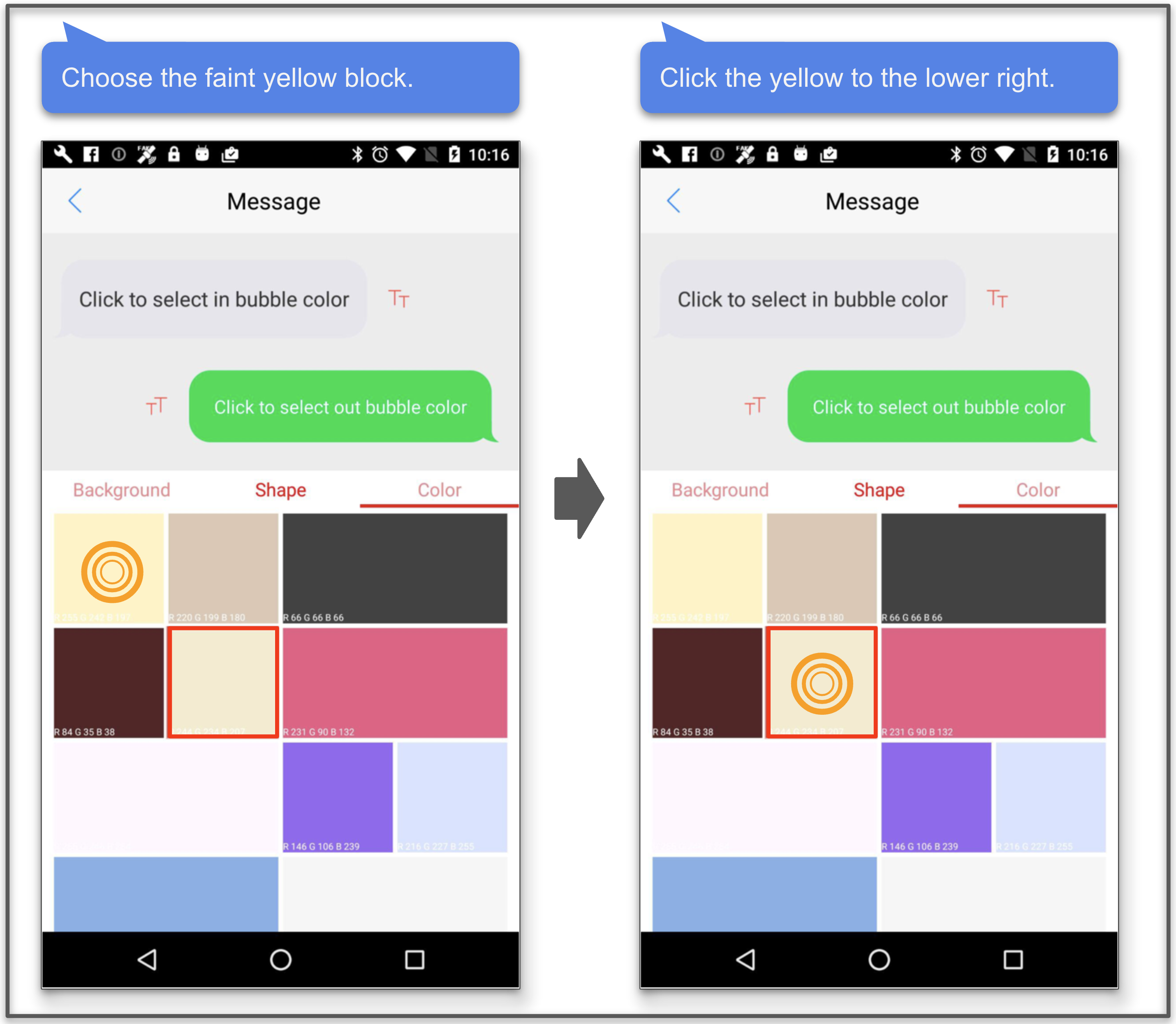}
    \caption{\small{Example two.}}
    \label{fig:intro_ex2}
  \end{subfigure}%
}
\caption{\small{Two illustrative examples of the \myname task. There are two turns in each of these examples. Interactions happen within a single screen. User commands are shown above the screens. The target object is bounded in \protect\inlinegraphics{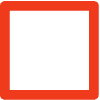}. Agent choices are marked with \protect\inlinegraphics{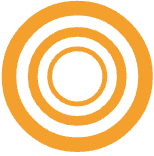}.}}
\label{fig:intro_example}
\end{figure*}

%In this paper, we investigate interactive grounding on GUIs, which involves multimodal input,
%, i.e., visual structures and image representation of a screen,
%and the action space is uniquely determined by interactive objects on a screen.
In this paper, we investigate interactive grounding on GUIs, which aligns multimodal input to actionable objects of a screen.
We focus on single-screen interaction which is the building block of UI reasoning.
%Although an interaction task typically involves multiple screens, our work focuses on command grounding on a single screen, which is a building block for multi-screen interaction. %To characterize interaction in the multi-modal setting, we focus on user interfaces (UI).
Specifically, we introduce the \myname (\textbf{M}ulti-turn \textbf{U}I \textbf{G}rounding) task in which the user iteratively guides the agent to select a desired UI object (see Fig.~\ref{fig:intro_example}).
With a given UI and a target object, the user instructs the agent via natural language, ranging from casual intent to more descriptive commands. %, e.g., \textit{"Mute direct message"} (see the 1st screen in Figure~\ref{fig:intro_example}).
The agent infers which UI object is intended by the user and and highlights it.
%If the highligh is correct, the user confirm the selection and the action is committed.
If the agent is correct, the user can confirm the selection and the grounding is completed.
%Otherwise the user issues another command, e.g., \textit{"Click the one below"}, to the agent to revise the selection (see the 2nd screen in Figure~\ref{fig:intro_example}). 
Otherwise, the user issues further guidance, e.g., \textit{"Click the one below"}, to the agent to refine its selection.
We collecte the \myname dataset from live interaction sessions between pairs of human annotators---one acts as the user and the other as the agent.
Our dataset has $77,820$ examples, each records the transaction history in a session.
%where an UI screen is shown to both the user and the agent yet a specified target object is only shown to the user such that the agent has to guess what the target object is from the user commands.
Specially, $20\%$ of the dataset are challenging ones as their human commands need multiple rounds to ground, even for human agents.

To establish the benchmark, we experiment with a range of variants to model the dynamics between the two roles. While the main goal of the task is to develop agent models for grounding, we also develop the user models for online instruction simulation.
%such that it can be used as a simulator to enable automatic online evaluation for interactive grounding.
We build our models upon a Transformer-based encoder-decoder architecture~\citep{Li2021-re}, and experiment with various learning methods, including traditional sequence modeling and reinforcement learning.
To fully examine the model performances, we evaluate the agent model with a spectrum of evaluation strategies, including both offline and online evaluations.
For the online evaluation, we employ both automatic and human evaluations, which include
interactions between the agent and the user (either a human or the user model) and offer a comprehensive probe into model understanding. %Neural models are known to suffer from potential artifacts in training data~\citep{ribeiro-etal-2020-beyond}, thus whether their predictions reflect \emph{true} understanding is often questionable. Being able to correct a model's own prediction error reflects a more comprehensive and robust understanding on the example~\citep{kassner-etal-2021-beliefbank}, and opens up more collaborative modeling options~\citep{schick2022peer, mishra-etal-2022-reframing}.
%For the agent, we explore non-interactive baseline, partially interactive one, and sequence models.
%For the user, we devise a simple and deterministic model based on template instantiation, as well as a neural version that is trained from human instructions.
%We use our user models to evaluate the agents extensively, 
Our experiments show that incorporating interaction substantially improves UI grounding task completion by $18\%$ on the entire dataset and $31\%$ on the challenging set, both in absolute scales.
Furthermore, our robustness measurements suggest \myname, while being a seemingly easy single-screen task, is actually difficult since neural agents sometimes struggle to correct themselves, resulting in repeated wrong selections across multiple turns.
This suggests large rooms for future improvement in grounding agents.

%Beyond accuracy, we also design a simple robustness metric for the agent, observing that neural agents still struggle to correct themselves as they often emit the same error ($\textt{\geq}37\%$) across multiple turns.

In summary, our key contributions\footnote{The dataset and code for reproducing our experiments will be released at http://github.com/TBD.} are:
\begin{enumerate} %[nosep]
\item We introduce \myname, a novel interactive vision-language task that focuses on multi-turn language grounding on a graphical UI screen, which is a challenging task that is meant to improve language grounding in realistic UIs.
\item We create a rich dataset that includes 77,820 examples recorded from live sessions between pairs of human users and agents. %which 20\% of the examples demonstrate highly ambiguous user commands and involve more than one round of interaction to complete.
And 20\% of the data are challenging for both human annotators and neural agents.
%\item We design two user models, one gives well-controlled and deterministic instructions, and the other is a neural version. Both effectively guide the agents towards task completion.
\item We experiment with a range of model variants and evaluation strategies, showing that iterative interaction significantly improves grounding accuracy by $18\%$ and $31\%$ on the entire and challenging test sets respectively, with automatic assistance from our user models. Our work lays a solid foundation for future investigations on collaborative grounding.
\end{enumerate}

%% file: background.tex
\section{Background}
%Grounding natural language in interactive/multi-modal setting has a long history of
Multi-modal modeling has a long history of research~\citep[e.g.,][]{winograd1972shrdlu,barnard2001learning,lavrenko2003model,Plummer2015-nl,Yu2016-am}.
One important area focuses on grounding objects in images where the natural language is used as an additional input \citep{Chen2017-fa, Yu2016-am, Yu2018-lg, FukuiPYRDR16, TransVG}.
%Instead of all the objects, only the ones referred by the natural languages are detected. %by predicting their bounding boxes.
%In this section, we focus on interactive and UI grounding.

\paragraph{Interactive Multimodal Grounding}
Prior works have formulated grounding as a multi-step reasoning task, e.g., navigation via multiple steps of grounding~\citep[e.g.,][]{ku-etal-2020-room, gur2018learning}.
Our work differs by focusing on agent's ability to self-correct in synchronized turns of interaction on a UI screen.
It is also conceptually linked to repeated reference game~\citep{hawkins2020characterizing}, except we use a different form of communication (language-action) instead of dialogue (language-language).
Our task leverages iteratively refined instructions on atomic action instead of the increased instruction utility over multi-step actions~\citep{effenberger-etal-2021-analysis-language}.
Our work models both the user and the agent, and let them communicate online.
This is different from single-sided modelings~\citep{suhr-etal-2019-executing, kojima-etal-2021-continual}.
Our observation that interaction improves grounding is also in line with dialogue-based works~\citep[e.g.,][]{haber-etal-2019-photobook, takmaz-etal-2020-refer}.

%On the other hand, interactively evaluating multimodal grounding agents offers more comprehensive probe on model understanding. Neural models are known to suffer from potential artifacts in training data~\citep{ribeiro-etal-2020-beyond}, thus whether their predictions reflect \emph{true} understanding is often questionable. Being able to correct a model's own prediction error reflects a more comprehensive and robust understanding on the example~\citep{kassner-etal-2021-beliefbank}, and opens up more collaborative modeling options~\citep{schick2022peer, mishra-etal-2022-reframing}.

\paragraph{UI Grounding}
%Object grounding for natural language has also been studied in the UI domain.
Grounding UI objects involves automatic completion of actions on web or mobile interfaces~\citep[e.g.][]{Pasupat2018-pg, Li2020-gl, He2020-nq}. %, which are critical for accessibility purposes.
It is also an important accessibility task for users who are situationally impaired when they are occupied by real-world tasks at hand \citep{Sarsenbayeva2018-ux}. 
Compared to grounding on natural images, these tasks usually take well-specified user commands
%(e.g., ``click the login button''),
and aim to select the object that best matches the command.
The UI image is often encoded via ResNet \citep{He2016-zd} or ViT \citep{Dosovitskiy2020-ds}. The structure and text features of UI are often encoded by Transformer model \citep{Vaswani2017-jm}. %, by flattening the objects in the view tree as input.
Fusing multimodal information is widely handled by cross-attention \citep[e.g.][]{He2020-nq, Li2021-re, bai2021uibert}.
We adopt these neural components in our benchmark.
%Finally, a matching score between an object and the command can be computed for ranking. 

%\paragraph{One-pass UI Grounding} Although a few grounding tasks have been proposed for user interfaces \cite{Pasupat2018-pg, Li2020-gl, itertask}, the problem is only considered with a single command or task description. We argue that it is more realistic to incorporate user interactions in the model training and evaluation. 
%User can
%give insufficient information in a single command or sometimes
%sometimes find it difficult to articulate the intention as the object might not be straightforward to describe.
%Based on the model prediction, the user can adjust the interface action in the follow-up command.
%Therefore, we propose a new grounding task based on multi-turn interactions. Offline reinforcement learning algorithms can be naturally applied to this task as the collected interactions can be viewed as expert demonstrations, and our goal is to train an agent model that can select the correct target object based on the commands.
%One immediate choice for modeling interaction is reinforcement learning.
%In this regard, we explore  for grounding multi-turn instructions.
%Extensive experiments show that the user interactions beyond single-shot command significantly improve the model performance.

%\paragraph{Connection to Reinforcement Learning}
\paragraph{Mobile UI Datasets}
Many grounding tasks, while covering multiple screens, remain one-pass reasoning, such as \textsc{PixelHelp}~\citep{Li2020-gl} and \textsc{MoTIF}~\citep{burns2022motifvln}.
Prior works~\citep[e.g.,][]{todi2021adapting} model sequence of (action, state) pairs via reinforcement learning (RL). 
%For instance, the agent responds with a sequence of actions to fulfill a function which spans over multiple interface states.
In contrast, \myname focuses on correcting a single action on one screen.
%Our task is not tied to a specific modeling approach.
%To show promising direction, we explore existing RL methods, i.e., Imitation Learning \citep{Pomerleau1989-gf} and Decision Transformer \citep{chen2021decision}.
%Compared to multi-screen tasks, we consider our approach allows the user to interactively interfere the execution of action sequences, thus complement traditional approaches.
Tab.~\ref{tab:data_comparison} summarizes key differences among other Mobile UI datasets.
Importantly, \myname is a challenging task as it enables corrective interaction in synchronized turn between user and agent.

\begin{table}[h]
\small
\setlength{\tabcolsep}{5pt}
\def\arraystretch{0.8}
    \centering
   \begin{tabular}{l|c|c|c|c}
     \toprule
     Data & Screen & Instruction & Natural Instruction & Interactive Correction \\
     \midrule
     \textsc{Rico}~\citep{Biplab-qr}  & multi & \xmark & \xmark & \xmark \\
     \textsc{PixelHelp}~\citep{Li2020-gl}  & multi & \cmark & \cmark & \xmark\\
     \textsc{MoTIF}~\citep{burns2022motifvln}  & multi & \cmark & \cmark & \xmark \\
     \textsc{RicoSCA}~\citep{Li2020-gl}  & single  & \cmark & \xmark & \xmark \\
     \textsc{RefExp}~\citep{bai2021uibert}  & single & \cmark & \cmark & \xmark \\
     \midrule
     \myname (Ours)  & single & \cmark & \cmark & \cmark \\
     \bottomrule
   \end{tabular}
   \caption{\small{Comparison to prior mobile UI Datasets.}}~\label{tab:data_comparison}
 \end{table}

%% file: task.tex
\section{Task Formulations} \label{sec:task}
As a grounding task, \myname involves two participants: a user and an agent. Our formulation includes both roles to provide a holistic view of interactive grounding.
The user's goal is to instruct, via natural language, the agent to select the desired object $g$ on the UI screen $S$.
%An action is executed by the agent inferring and selecting a target element on the screen that corresponds to the action. 
The unique aspect of \myname is that it allows the user to guide the agent \emph{iteratively} to identify the target action by issuing a series of commands, each in response to the agent's prior inference. 

We separate such user-agent interaction into turns. At turn $t$, the interaction consists of:
\begin{align*}
    \begin{cases}
      c_t: \text{user command},\\
      a_t: \text{agent action}.
    \end{cases}
\end{align*}
where the user first instructs the agent with command $c_t$, and the agent responds with a suggestion of action $a_t$.
Here $a_t$ is essentially the index of object. The task is completed when $a_t=g$.

\subsection{Agent Task}
In \myname, the action space for the agent consists of a set of UI objects to click on the interface, e.g., in Fig~\ref{fig:intro_example}.
%In practice, we disable long-press or and multi-screen functions such as scrolling.
%shows the context in the middle of the third turn.
%In the first turn, $c_0=$ \emph{``Want to focus on my work''} and the agent acted with $a_0=$(2).
%In the second turn, the user issued a correction $c_1=$ \emph{``Move to the above''},
%and the agent responded $a_1=$(4).
%Note that, the agent can only respond by selecting visual objects.
%This differs from a typical conversation task where both user and agent can use language to communicate.
Intuitively, we would want the agent to take the desired action $g$ as early as possible, i.e., as few turns as possible.
Thus, at turn $t$, the agent models
\begin{align}
    P_\theta(a_t| S, c_{[0, t]}, a_{[0, t-1]})   \label{eq:agent_task}
\end{align}
where $\theta$ denotes the agent parameters.
This iterative grounding early stops once $a_t=g$ or $t$ reaches a maximum number of turns allowed.

%However, Eq~\ref{eq:evaluation} assumes user instructions are all correct and well-specified, while in practice they can often be ambiguous, especially when the visual structure is complicated.
%As we will see in Sec. \ref{sec:dataset}, users often give short and confusing commands, which are then corrected/clarified in follow-up turns.

%Therefore, in \myname, we also consider the agent's capability to complete the task by learning from multi-turn interaction, including error-correcting instructions.
%That is,
%\begin{align}
%    \foneat{\text{all}} = P (a_T = g | S, c_{[0, T]}, a_{[0, T)}) \label{eq:evaluation_all}
%\end{align}
%where $T$ denotes the last turn for the given example.
%Note that, since we compute F1 with early stop, the last turn would have the correct selection from the agent.
%The task also stops after a number of turns.

%In an extreme case, we consider an agent with high \foneat{0} but incapable of refining its own prediction to be problematic, since it questions the agent's understanding on the interface.
%Similarly, an agent that is insensitive to corrective instructions is also problematic.
%The latter case is essentially a robustness measurement
%\begin{align}
%    \gammaat{t} = P(|\{a_{[0, t]}\}| \neq t) \label{eq:gammadefinition}
%\end{align}
%which is the percentage of examples that have duplicate actions within $t$ turns.
%In this paper, we will take a joint consideration of Eq.~\ref{eq:evaluation}\&\ref{eq:evaluation_all} when evaluating agent performances.

\subsection{User Task} \label{sec:user_task}
The user's role is to provide guidance to the agent through iteratively refined instructions.
In contrast to one-pass prediction tasks~\citep[e.g.][]{Pasupat2018-pg, He2020-nq} where the agent makes a one-shot guess,
a \myname user issues follow-up commands that are dependent of prior instructions $c_{[0,t-1]}$ and agent actions $a_{[0, t-1]}$,
%The role of user is to provide sufficient guidance to the agent towards $g$.
%In one-time prediction task, the user would articulate the action $g$ in details such that the agent has enough information to reason with.
%In contrast, \myname users complete the task via iteratively refined instructions.
%In this paper, we focus on the latter case as it befits realistic use cases.
%Correcting agent actions assumes the first instruction $c_0$ is already given.
%Before turn $t$, if the agent has not selected $g$, a follow-up instruction is issued to:
%\begin{itemize}[noitemsep]
%    \item correct the prior agent selection $a_{t-1}$;
%    \item clarify prior instructions $c_{[0,t)}$;
%    \item correct prior instructions $c_{[0,t)}$.
%\end{itemize}
which is formalized as the following:
\begin{align}
    P_\phi(c_t|S, g, c_{[0, t-1]}, a_{[0, t-1]})
    \label{eq:user_model}
\end{align}
where $\phi$ denotes the user. Here, the user model is aware of the target object $g$.
%Note that Equation~\ref{eq:user_model} formalizes the task of the user simulator model that we develop for automatic online evaluation.

\paragraph{Interplay between User and Agent} The agent task (Eq.~\ref{eq:agent_task}) is the pivot of \myname. The user task (Eq.~\ref{eq:user_model}) aims to guide agent towards task completion, which potentially includes online training.
In our benchmark, we let the user and agent play together.
Although automatic evaluation is not as realistic as human evaluation, it offers a fast, low-cost, and reproducible environment. 
This setting also allows us to study various questions surrounding the interplay between the two, e.g., whether an automatic user can assist an agent?
and whether agent errors would confuse the user?
%A user model can be used to evaluate the agent model in an online fashion, and potentially train the agent on the fly.
%This gives a chance for degraded performance for both user and agent, e.g., both agent and user repeating the same prediction. In Sec.~\ref{sec:experiments}, we will take joint consideration on such cases. 

%\tao{Talk about the interplay btw agent and user, and cite Sebastian's paper.}

%The evaluation of user instruction is two-folded.
%One one hand, we care about the downstream impact on the agent performances.
%One the other hand, we want an user model to generate instructions aligning well to those from human.
%The former can be reflected by improvements in Eq.~\ref{eq:f1definition}\&\ref{eq:gammadefinition}.
%For the latter, we will use CIDEr~\citep{vedantam2015cider} for salable quantitative measurement.
%In Sec.~\ref{sec:experiments}, we will jointly consider these two aspects.

%\tao{Also talk about we prefer both short path and completeness of the task}
%\paragraph{Connection to generative adversarial networks (GAN)}.
%Instead of competition, the participants in \myname are required to work collaboratively as it befits real use cases.

%% file: data.tex
\section{Dataset Creation}\label{sec:dataset}

As there is no available dataset for model training and evaluation, we developed an interactive labeling interface to collect data for \myname.
Our data collection involves two human annotaters to play the roles of the user and the agent respectively in a live session.
The user and the agent have two separate views, running on different machines (Appx.~\ref{sec:labeling_interface}).
Both views share the same UI screen and a message box showing instruction history.
Our task embodies the \emph{eyes-on, hands-free} situation for mobile interaction where the user is required to only use language for the task, and the machine responds its prediction by highlighting.
The user can commit the action if the prediction is confirmed.
During a session, only the user can see the target; and the the message box is read-only to the agent thus no language-base dialogue would take place.
%Thus, several differences between user/agent views are:
%1) by design, only user can see the target object;
%2) message box is read-only to the agent thus no language-based dialogue.
%But there are several critical differences between the user and the agent's view. 
%The target object is highlighted in the user's view but not in the agent's view. It is designed in such a way that the user can compose the language command based on the given target on the user view, and the agent can only guess what the intended target is based on the user command. While the user can enter text in the message box, the message box is read-only for the agent whose response is only through selecting objects on the screen.

%To stay focused on single-screen interaction, we only allow clickable objects visible on the current screen.
%Multi-screen functions are disallowed.

\subsection{Annotation Workflow}
We use the UI corpus, mobile UI screenshots and view hierarchies, from \textsc{Rico}~\citep{Biplab-qr} and auxiliary object features from the \textsc{Clay} dataset~\citep{clay}. Each session starts with a randomly sampled UI object (e.g., a \emph{button}), from the visible view hierarchy, as the target object $g$.
%We filter out container-like types (e.g., a \emph{card\_view}) in the view hierarchy as they are less likely to be clicked compared in real use cases.
%With the $g$ given, the user issues an initial command $c_0$.
User annotators are encouraged to articulate their initial command ($c_0$) casually or goal-oriented.
%, to demand
%intelligent interactive behaviors
%communication from the agent.
%This design of annotation poses challenges to modeling, while it addresses the situation where visual objects are difficult to articulate (e.g., no text labels), and potentially reduces the effort for the user to compose precise and literal commands in realistic scenarios.
We consider such design to cover the realistic scenarios discussed in Sec.~\ref{sec:intro},
and potentially free users from composing long and precise instructions.

%Through our pilot studies, we recognize two benefits of goal-driven $c_0$: 1) it covers scenarios where visual objects are difficult to articulate (e.g., no explicit text/function); 2) it increases annotation efficiency as it promotes more turns of interaction. visual objects are difficult to articulate (e.g., no explicit text/function)

In the agent view, all clickable objects on the UI screen are revealed with their bounding boxes highlighted, which show what objects the agent can select, without indicating which one is the target $g$.
The current agent selection is reflected on both the user and the agent's view. 
%It is possible that multiple objects have their bounding boxes occlude each other.
%In this case, we allow agents to loop object selection by clicking the same position multiple times.
%The selection is only visible to the user after it is manually submitted.
The session continues to the user's turn if the agent selection does not match $g$.
%in which case the user issues a follow-up command.
In follow-up turns, the user is not allowed to repeat a command issued in previous turns, and likewise the agent is not allowed to select an previously chosen object.
Upon the agent selection matching the target object in the user view the task is completed.
Each session allows up to $5$ turns and we filter out those unfinished.

%\tao{Put a 2-column fig for a complete example.}
%\tao{Where does our screen come from? How did we split the data? (random sampling?) Filtering/post processing?}

%To build the initial dataset for labeling, we use mobile UI screenshots and view hierarchies from the RICO dataset \cite{Biplab-qr} and interface object labels from the Clay dataset \cite{clay}. We randomly sample leaf objects (e.g., \emph{button}) with non-container types, and use these objects as the target objects for the user-agent interaction. We believe these objects are more likely to be clicked compared to the container types (e.g., \emph{card\_view}). 
%The users are asked to send instructions focusing on the target object and interaction history, without repeating itself.
%For each pair of (interface $S$, target object $g$), we collect one track of interaction history\footnote{\myname instructions are in English, to be consistent with the language in most of our interfaces.}.

%The user is instructed to issue commands to select the target object, while the agent tries to click on the target object based on the commands. The user is asked to send commands focusing on the target object, but not random commands for open exploration of the UI. We collect one interaction for each example.

\subsection{Results \& Analysis}

\begin{table*}[h]
\small
\def\arraystretch{1}
  \centering
  \scalebox{0.9}{
  \begin{tabular}{l|r|r|r|c|c|c|c|c|c|c}
    \toprule
    \multicolumn{6}{c}{Statistics of examples} & \multicolumn{5}{|c}{Distribution of Turns (\%)} \\
    \midrule
    {Split} & {Apps} & {Screens} & {Interactions} & {Avg. \#Turns} & {Avg. \#Token/Turn} & {1} & {2} & {3} & {4} & {5} \\
    \midrule
    Train & 6,039 & 26,090 & 65,235 & 1.24 & 4.26 & 78.91 & 18.31 & 2.37 & 0.35 & 0.06 \\
    Dev & 544 & 2,625 & 6,377 & 1.23 & 4.18 & 79.99 & 17.77 & 1.91 & 0.27 & 0.06 \\
    Test & 549 & 2,550 & 6,208 & 1.23 & 4.18 & 80.20 & 16.82 & 2.55 & 0.40 & 0.03 \\
    \midrule
    All & 7,132 & 31,265 & 77,820 & 1.24 & 4.25 & 79.10 & 18.15 & 2.35 & 0.35 & 0.06 \\
    \bottomrule
  \end{tabular}}
  \caption{\small{Dataset statistics.}}~\label{tab:dataset}
\end{table*}

We collected 77,820 examples based on 31,265 unique screens from 7,132 apps (see details %statistics of our dataset
in Table~\ref{tab:dataset}).
We split the dataset into the training, development, and test sets.
We use app-wise split~\citep{Li2020-la} to avoid potential leaking across sets.
%such that no screens from the same app are shared by these splits to avoid potential information leak.
%\tao{mention that we only take one path for each example, no open exploration, also bring this up in the limitations section}. \gang{added at the end of the previous paragraph}
%\tao{have we talked about any filtering?} \gang{filtering briefly mentioned in the previous paragraph}
As shown in Table~\ref{tab:dataset}, the three splits have a similar distribution regarding the number of turns in each example.

In general, users tend to provide short instructions ($\sim4$ words).
Across the dataset, annotators completed $\sim80\%$ examples in one turn (
%using a single user command and a single agent selection
i.e., one instruction and one selection
).
The rest $20\%$ dataset is considered to be more challenging, with the majority taking $2-3$ turns.
We call this $20\%$ as the \emph{Challenging} subset.
In Sec.~\ref{sec:experiments}, we will show that examples requiring more turns for human agents are also more difficult for grounding models.
We refer readers to Appx.~\ref{sec:mug_examples} for examples.

Note that while the single-turn cases are relatively easy for humans, they are important for establishing a baseline of understanding on the task.
In online evaluations (Sec.~\ref{sec:experiments}), we will see that models could also struggle with a nontrivial portion of these single-turn cases, which indicates that agents can benefit from interactions with the user to correct their mistakes.

In Table~\ref{tab:challenging_analysis}, we categorize $200$ Challenging examples from the development split.
We found follow-up commands are mainly to make spatial adjustments ($50\%+10\%$) or to add extra information.

%\tao{what is our take on this 20\% examples?} -- added analysis below

\begin{table}[h]
\small
\def\arraystretch{0.9}
    \centering
   \begin{tabular}{r|l|l}
     \toprule
     Percentage & Attribution & Example \\
     \midrule
     50\% & Adjusting relative position in the layout. & \emph{the value before the text.}\\
     31\% & Providing more information of the target. & \emph{show me channels.} $\rightarrow$ \emph{click tv icon.} \\
     10\% & Adjusting direction/position on the screen. & \emph{not reward but collect at the bottom.} \\
     3\% & Rephrasing the instruction. & \emph{go to books.} $\rightarrow$ \emph{show me books logo.} \\
     \bottomrule
   \end{tabular}
   \caption{\small{Major categories for the second turn from $200$ examples in the development split.}}~\label{tab:challenging_analysis}
 \end{table}

%% file: modeling.tex
\section{Grounding Models}\label{sec:modeling}
Our agent model is based on a transformer encoder-decoder architecture. The encoder comprehends the objects on the UI screen (Sec.~\ref{sec:ui_encoder}) and the decoder predicts a target action based on the object encodings, the user commands, as well as previous predictions (Sec.~\ref{sec:grounding_decoder}).
%Fig.~\tao{need a 2-colume model illustration} illustrates an overview of our agent model.
%Finally, we will extend the encoder-decoder architecture to design a user model (in Sec.~\ref{sec:user_model}).

\subsection{UI Encoder}
\label{sec:ui_encoder}
Our encoder processes the interface $S$.
%Our encoder takes multimodal features from the UI.
Each $S$ consists of two modalities of information, i.e., a screenshot $I_S$ and view hierarchy features $\psi$~\citep{Biplab-qr, clay}. The concrete list of $\psi$ is in Appx.~\ref{sec:vh_features}.
%We will refer $\psi^k$ as features for the $k$-th object in $S$.
The output is an encoding $v^k$ for each object indexed by $k$, similar to~\cite[e.g.,][]{Li2020-gl, He2020-nq, Li2020-la}:
\begin{align}
    \Phi_S &= \text{ResNet}(I_S) \\
    v &= T_\text{enc}(\{ \text{ROI}^k(\Phi_S) | \psi^k \}) \label{eq:encoder}
\end{align}
For the image, we use a pre-trained ResNet-50~\citep{He2016-zd} which is fine-tuned with other modules.
%and fine-tuned on the \myname training set.
The resulted $\Phi_S$ (grid size of $h\times w$)
is then mapped to object level by region-of-interest (ROI) pooling~\citep{RenHG015}.
Finally, we fuse multimodal features for each object by a transformer encoder $T_\text{enc}$.
The output $v$ stands for a sequence of object encodings which are interaction-agnostic.
%We next join the image encoding and the view hierarchy features of each object using concatenation, which is fed to a transformer encoder
%, to align image and view hierarchy features $\psi$, where $r$ and $\psi$ are concatenated object-wise%, and the sequence is arranged by their pre-order traversal in the view hierarchy.
%At this point, $v^k$ is the interaction-agnostic encoding for the $k$-th object.

\subsection{Grounding Decoder} \label{sec:grounding_decoder}
We use a causal transformer $T_\text{dec}$ to predict from interaction history.
The architecture is similar to~\citep{Li2021-re}, except we use multi-turn interaction as input, e.g., $c_{[0,t]}$ and $a_{[0,t-1]}$.
The output of $T_\text{dec}$ is a vector $z_t$ that summarizes prior interaction up to $c_t$:
\begin{align}
    z_t = T_\text{dec} (v, c_0, v^{a_0}, c_1, ..., v^{a_{t-1}}, c_t) \label{eq:grounding_decoder}
\end{align}
where $a_t$ denotes object index, either from model prediction or human selection.
%where $v^{a_{t-1}}$ denotes the encoding for the $a_{t-1}$-th object (see Sec.~\ref{sec:ui_encoder}) which could be either the object of model prediction or human selection.
The specific input to Eq.~\ref{eq:grounding_decoder} will be subject to modeling variants in Sec~\ref{sec:agent_models}.
%Moreover, Eq.~\ref{eq:grounding_decoder} is subject to specific modeling variant we will see in Sec~\ref{sec:agent_models}.
For classification, we use a linear layer $f$ to score the $k$-th object:
\begin{align}
    a_t = \arg\max_k f([z_t|v^k])
\end{align}

%\paragraph{Variant for Imitation Learning}
%Note that the model in Eq.~\ref{eq:grounding_decoder} is directly compatible with imitation learning.
%During training, we can take all annotator selections within an example to supervise the model.

%\paragraph{Variant for Offline RL}
%For RL modeling, we can extend Eq.~\ref{eq:grounding_decoder} following the setup in decision transformer~\cite{chen2021decision}:
%\begin{align}
%    z_t = T_\text{dec} (\{v^k\}; c_0; w_0; v^{a_0}; c_1; ...; v^{a_{t-1}}; c_t; w_t) \label{eq:rl_decoder}
%\end{align}
%where $w_t$ denotes the return token from turn $t$ to the end of the task.

%\subsection{Instruction Decoder} \label{sec:user_model}
%Recall in Sec.~\ref{sec:task}, the goal for user is to give corrective guidance to the agent.
%To model this user behavior, we can use a similar decoder module to generate corrective instructions:
%\begin{align}
%    c_t = T^{\text{greedy}}_\text{dec} (v; c_0; v^{a_0}; c_1; ...; v^{a_{t-1}};) \label{eq:ins_decoder}
%\end{align}
%where $T^{\text{greedy}}_\text{dec}$ denotes $T_\text{dec}$ followed by greedy decoding.

%% file: experiments.tex
\section{Experiments} \label{sec:experiments}
The goal of our experiments is to explore training and evaluation methods for \myname and establish a benchmark.
In this section, we discuss training and evaluation setups for the architecture discussed in Sec.~\ref{sec:modeling}.
We will explore multiple variants for the agent in Sec.~\ref{sec:agent_models}.
For automatic evaluation, we present a simple and effective heuristics-based user model and a neural version in Sec.~\ref{sec:user_models},.
Finally, we show extensive F1 results in Sec.~\ref{sec:offline_results} and \ref{sec:online_results}, robustness in~\ref{sec:agent_robustness}, ablations in~\ref{sec:auto_human_eval} and \ref{sec:ablation_instruction}, and error analysis in~\ref{sec:error_analysis}.
We refer readers to Appx.~\ref{sec:hyperparameters} for hyperparameters and~\ref{sec:pred_examples} for model predictions.

%We consider a broad spectrum of modeling options for the task, using the $T_\text{enc}$ and $T_\text{dec}$ (in Sec.~\ref{sec:modeling}) as a backbone.
%To stay focused, we experiment with immediate variants to the general encoder-decoder framework.
%We also propose user models for automatic and scalable online evaluation.
%Our agent models are all trained offline to avoid potential overfitting to the user models.
%We refer readers to Appendix~\ref{sec:hyperparameters} for hypeparameters.

To avoid confusion, we thereafter use $a_t^\prime$ to refer to the selection predicted by the agent model at turn $t$, while $a_t$ to the human agent's selection.
Similarly, we refer $c_t^\prime$ to instruction generated by the user model while $c_t$ to the one from the human user.

\subsection{Agent Models} \label{sec:agent_models}
Our agent models use the $T_\text{enc}$ and $T_\text{dec}$ (in Sec.~\ref{sec:modeling}) as a backbone, denoted as $\theta$.
Recall that $T_\text{enc}$ processes $S$ while $T_\text{dec}$ processes interaction.
Here, we discuss different handlings of $T_\text{dec}$.
%We focus on the agent side, comparing models training on different aspects of the data to show the improved grounding from using multi-turn interaction.
%Across different models below, we use their offline validation performance defined in Eq.~\ref{eq:evaluation_all} for model selection.
%To differentiate notation, we will refer $a_t^\prime$ to model predicted action at turn $t$, while $a_t$ to the annotator's selection.
%We explore different variants to our interactive agent model, including imitation and offline RL model.

\paragraph{Single or Multi-turn Model}
The first factor we investigate is how allowing multiple turns helps grounding. For each example, we can feed the entire interaction history as input to the agent model and supervise agent selection on the very last turn $T$:
\begin{align}
    P_\theta (a_T^\prime = g | S, c_{[0, T]}, a_{[0, T-1]}) \label{eq:multiturn_training}
\end{align}
%where $\theta$ denotes the agent model (i.e. $T_\text{enc}$ and $T_\text{dec}$ in Sec.~\ref{sec:modeling}).
%Alternatively, we can also supervise at every intermediate turn instead of only the last one.
%However, models trained this way performed similar to Eq.~\ref{eq:multiturn_training} in offline measurements.
We can further reduce the input to be $(S, c_0)$ only, making a single-turn model.
%\begin{align}
%    \arg\max_\theta P_\theta (a_0^\prime = g | c_0) \label{eq:singleturn_training}
%\end{align}
To evaluate single-turn model with multi-turn examples, we simply concatenate all $c_t$ into one instruction.

\paragraph{Instruction-only Model}
To understand how it helps grounding by taking into account of previous actions of the agent in the multi-turn model (Eq.~\ref{eq:multiturn_training}), we introduce the command-only baseline, which ignores agent actions (selections) in the interaction history:
%To have ablation study on whether agent selections $a_{<T}$ help, we can isolate them and have an instruction-only agent:
\begin{align}
    P_\theta (a_T^\prime = g | S, c_{[0, T]}) \label{eq:instruction_only_training}
\end{align}
%Note the difference from Eq.~\ref{eq:multiturn_training} is that this model ignores annotator selections in between.
%This model avoids incorrect selections in between but also misses part of the interaction.

\paragraph{Imitation Model}
Instead of supervising the agent only at the last turn, we can model the entire action sequence as an imitation model:
\begin{align}
    \prod_t P_\theta (a_t^\prime = a_t | S, c_{[0, t]}, a_{[0, t-1]}) \label{eq:imitation_training}
\end{align}
This variant investigates whether the supervision of the intermediate actions helps.
%Alternatively, we can also focus on early completion of the agent task by replacing $a_t$ with $g$ for each turn.
%However, in our preliminary experiments, we found such model performed similar to our multi-turn model in Eq.~\ref{eq:multiturn_training}.
%This suggests that, with a substantial amount of confusing instructions need further clarification, learning to early complete involves too much noise.

\paragraph{Offline RL}
%Another immediate sequence modeling option is RL.
Lastly, because each turn the agent action affects how the user responds, \myname can be formulated as a RL problem where  the user and the UI constitute the environment.
We use the Decision Transformer~\citep{chen2021decision} for offline RL, by inserting extra learnable return tokens $w_t$ to the $T_\text{dec}$ before each action: $T_\text{dec}(v, c_0, w_0, v^{a_0},...,c_t, w_t)$. The model is:
\begin{align}
    \prod_t P_\theta (a_t^\prime = a_t | S, c_{[0, t]}, w_{[0, t]}, a_{[0, t-1]}) \label{eq:offline_rl}
\end{align}
%where $w_t$ denotes the reward token embedding for turn $t$.
Possible discrete return tokens are $\{1, 2, 3, 4\}$ where $1$ on the last turn.
%Possible reward tokens in our problem are $\{1, 2, 3, 4\}$.
%Our training follows~\citet{chen2021decision} to have reward $1$ on the last turn and iteratively increment in prior turns.
During testing, we follow~\citet{chen2021decision} to force the current turn to have return $1$ and adjust prior returns.
%and increment prior rewards.

\subsection{User models} \label{sec:user_models}
Here, we design a simple and effective heuristics-based user model, and then develop a neural version.
To show automatic online evaluation is a promising direction for \myname, we also conducted human evaluation on a shared set of $500$ examples from the test split (Sec.~\ref{sec:auto_human_eval}).

\paragraph{Heuristics-based Model}
We observe that, when the selection $a^\prime$ is incorrect, we can deterministically devise a follow-up instruction by using a template as below:
\begin{align*}
    \text{Not the \red{$a_t^\prime$}, click the \gold{$g$} to/on the \blue{dir}.}
\end{align*}
This template is to be instantiated on view hierarchy features (in Appx.~\ref{sec:vh_features}).
Compared to human follow-ups, heuristic ones are more specific and longer, such as:
{
\small
\begin{itemize}[noitemsep]
    \item Not the \red{icon}, click the \gold{action notifications} on the \blue{top right of the screen}.
    \item Not the \red{text}, click the \gold{input search} to the \blue{slight right and below of your choice}.
\end{itemize}
}

\paragraph{Neural Instruction Model}
We extend the \emph{Multi} agent architecture to model follow-up commands:
%We modify the \emph{Multi} agent to generate follow-up instructions:
\begin{align}
    P_\phi (c_t^\prime = c_t | S, g, c_{[0, t-1]}, a_{[0, T-1]}) \label{eq:user_training}
\end{align}
%\begin{align}
%    c_t = T^{\text{greedy}}_\text{dec} (v; c_0; v^{a_0}; c_1; ...; v^{a_{t-1}};) \label{eq:ins_decoder}
%\end{align}
%where $T^{\text{greedy}}_\text{dec}$ denotes a transformer decoder followed by greedy decoding.
which has decoder $T_\text{dec}(v, v^g, c_0, v^{a_0}, c_1, ..., v^{a_{t-1}})$ at turn $t$.
For training, we teacher-force at each turn ($t>0$).
%For training at turn $t$, we use interactions from annotators in prior turns as input, and teacher-force on $c_t$.
We found that using heuristics as prompt greatly boosts development CIDEr~\citep{vedantam2015cider} to from $70$ to $78$.
For inference, we use greedy decoding with a maximum length of $12$.

\subsection{Metrics} \label{sec:metrics}
We focus on evaluating the agent model as it is the pivot task of \myname.
Intuitively, we would want the agent to take the desired action $g$ with as few turns as possible.
That is,
\begin{align}
    \foneat{t} &= \sum_t P (a_t = g | S, c_{[0, t]}, a_{[0, t-1]}) \label{eq:f1definition}
\end{align}
where, in practice, we compute \foneat{t} with early stop over turns to avoid double counting.
Clearly, an agent with high F1 and a lower value of $t$ is better than an agent that requires more turns for the same accuracy.
%Indeed, when adapting Eq.~\ref{eq:agent_task} in model training, it implicitly minimizes the number of turns.
With $t$ limited to $0$, the task is reduced to a one-pass grounding task.

In an extreme case, we consider an agent with high \foneat{0} but flat changes in \foneat{$t>0$} to be problematic, since it questions the agent's understanding about the interface.
For more comprehensive testing, we also use a simple robustness metric for prediction changes across turns:
\begin{align}
    \Gamma = P(|\{a_{t}\}| \neq T) \label{eq:gammadefinition}
\end{align}
which is the percentage of examples that have duplicate actions within $T$ valid turns.
We observe that user models, unlike human annotators, sometimes give the same instruction across turns, especially when the agent repeats the same error.
Therefore, we only count agent selections that are associated with unique instructions in each example, and ignore other turns.

\subsection{Offline Results} \label{sec:offline_results}

\begin{table}[t]
\small
\setlength{\tabcolsep}{5pt}
\def\arraystretch{1}
  \centering
  \scalebox{0.9}{
  \begin{tabular}{l|c|c|c|c|c|c!{\vrule width 1pt}c|c|c|c|c|c}
    \toprule
    \multicolumn{7}{c!{\vrule width 1pt}}{Challenging} & \multicolumn{6}{c}{All}  \\
    \midrule
    {Model} & {\foneat{0}} & {\foneat{1}} & {\foneat{2}} & {\foneat{3}} & {\foneat{4}} & {avg$_\text{std}$} & {\foneat{0}} & {\foneat{1}} & {\foneat{2}} & {\foneat{3}} & {\foneat{4}} & {avg$_\text{std}$} \\
    \midrule
    Single & \tb{26.8} & 44.7 & 45.6 & 45.7 & 45.7 & 46.1$_{1.3}$  & 56.9 & 60.5 & 60.7 & 60.7 & 60.7 & 60.3$_{0.8}$ \\
    Ins-only & 25.2 & 49.7 & 52.1 & 52.2 & 52.2 & 53.5$_{1.3}$    & 58.5 & 63.4 & 63.8 & 63.8 & 63.9 & 64.0$_{0.5}$ \\
    Multi & 25.2 & 54.2 & 57.2 & 57.4 & 57.4 & \tb{59.9}$_{1.5}$  & \tb{58.6} & \tb{64.3} & \tb{64.9} & \tb{64.9} & \tb{64.9} & \tb{65.1}$_{0.2}$ \\
    Imitation & 23.5 & \tb{56.5} & \tb{59.6} & \tb{59.6} & \tb{59.6} & 59.4$_{1.5}$    & 56.6 & 63.1 & 63.7 & 63.7 & 63.7 & 64.0$_{0.8}$ \\
    Offline RL & 24.2 & 55.4 & 58.1 & 58.2 & 58.2 & 58.1$_{1.1}$     & 58.0 & 64.2 & 64.7 & 64.8 & 64.8 & \tb{65.1}$_{0.5}$\\
    \bottomrule
  \end{tabular}}
  \caption{\small{Offline agent F1$\uparrow$ on the test set. \foneat{0-4} are from model trained with seed 1 and avg$_\text{std}$ is \foneat{4} of 5 runs.}}~\label{tab:offline_evaluation}
\end{table}

Tab.~\ref{tab:offline_evaluation} presents offline results on the test set, over the \emph{Challenging} (see Sec.~\ref{sec:dataset}) and the \emph{All} sets.
During inference, we use instructions from the human user and actions from the human agent for turns in between and ask an agent model to predict at each turn.
%Again, as mentioned in Sec.~\ref{sec:intro}, once the agent has predicted correctly, the example is completed, thus no further turn is considered.
%The score \foneat{all} is computed using the complete interaction history for each example.
Doing so requires agent models to correct human agent actions, instead of the model's own.
Clearly, the models that take into account interaction history outperform those use none or partially.
While the \emph{Ins-only} and the \emph{Imitation} models perform closely on the \emph{All} set, the \emph{Imitation} is stronger on the \emph{Challenging}, and they bear larger margins in online tests.

%\begin{table}[ht]
%\small
%\setlength{\tabcolsep}{3pt}
%\def\arraystretch{1}
%  \centering
%  \scalebox{0.98}{
%  \begin{tabular}{l|c|c|c|c|c|c}
%    \toprule
%    \multicolumn{7}{c}{Challenging} \\
%    \midrule
%    {Model} & {\foneat{0}} & {\foneat{1}} & {\foneat{2}} & {\foneat{3}} & {\foneat{4}} & {\foneat{all}} \\
%    \midrule
%    Single & \tb{26.8} & 44.7 & 45.6 & 45.7 & 45.7 & 39.1 \\
%    Ins-only & 25.2 & 49.7 & 52.1 & 52.2 & 52.2 & 48.3 \\
%    Multi & 25.2 & 54.2 & 57.2 & 57.4 & 57.4 & \tb{52.0} \\
%    Imitation & 23.5 & \tb{56.5} & \tb{59.6} & \tb{59.6} & \tb{59.6} & 50.4 \\
%    RL & 24.2 & 55.4 & 58.1 & 58.2 & 58.2 & \tb{52.0} \\
%    \midrule
%    \multicolumn{7}{c}{All} \\
%    \midrule
%    Single & 56.9 & 60.5 & 60.7 & 60.7 & 60.7 & 59.4 \\
%    Ins-only & 58.5 & 63.4 & 63.8 & 63.8 & 63.9 & 63.1 \\
%    Multi & \tb{58.6} & \tb{64.3} & \tb{64.9} & \tb{64.9} & \tb{64.9} & \tb{63.9} \\
%    Imitation & 56.6 & 63.1 & 63.7 & 63.7 & 63.7 & 61.9\\
%    RL & 58.0 & 64.2 & 64.7 & 64.8 & 64.8 & 63.5 \\
%    \bottomrule
%  \end{tabular}}
%  \caption{Offline evaluation of agents on the test split.}~\label{tab:offline_evaluation}
%\end{table}

\begin{table*}[t]
\small
\setlength{\tabcolsep}{5pt}
\def\arraystretch{1}
  \centering
  \scalebox{0.9}{
  \begin{tabular}{c|l|c|c|c|c|c|c!{\vrule width 1pt}c|c|c|c|c|c}
    \toprule
    \multicolumn{8}{c!{\vrule width 1pt}}{Heuristics} & \multicolumn{6}{c}{Neural} \\
    \midrule
    & {Model} & {\foneat{0}} & {\foneat{1}} & {\foneat{2}} & {\foneat{3}} & {\foneat{4}} & {avg$_\text{std}$} & {\foneat{0}} & {\foneat{1}} & {\foneat{2}} & {\foneat{3}} & {\foneat{4}} & {avg$_\text{std}$} \\
    \midrule
    \multirow{5}{*}{{\rotatebox[origin=c]{90}{\footnotesize{Challenging}}}}  & Single & \tb{26.8} & 39.8 & 43.3 & 44.6 & 44.6 & 44.1$_{0.5}$ &    \tb{26.8} & 41.7 & 43.9 & 44.6 & 45.2 & 44.9$_{1.0}$ \\
    & Ins-only & 25.2 & 47.4 & 51.7 & 52.9 & 53.5 & 52.9$_{1.4}$ &     25.2 & 43.4 & 46.5 & 48.2 & 48.5 & 49.1$_{0.7}$ \\
    & Multi & 25.2 & \tb{47.8} & 50.9 & 51.7 & 52.4 & 54.3$_{1.1}$ &     25.2 & 43.9 & 47.4 & 48.9 & 49.4 & 50.0$_{1.1}$ \\
    & Imitation & 23.5 & 39.8 & 43.3 & 46.8 & 48.1 & \tb{55.2}$_{0.4}$ &      23.5 & 44.1 & \tb{51.4} & \tb{55.5} & \tb{57.6} & \tb{57.7}$_{1.5}$\\
    & Offline RL & 24.2 & 47.6 & \tb{52.7} & \tb{54.1} & \tb{54.6} & 54.6$_{1.2}$ &    24.2 & \tb{44.6} & 49.4 & 51.3 & 52.0 & 53.4$_{1.3}$ \\
    \midrule
    \multirow{5}{*}{{\rotatebox[origin=c]{90}{All}}} & Single & 56.9 & 65.2 & 67.4 & 68.1 & 68.1 & 68.7$_{0.8}$ &   56.9 & 65.0 & 66.5 & 67.0 & 67.4 & 67.1$_{0.8}$ \\
    & Ins-only & 58.5 & 70.9 & 72.9 & 73.6 & 74.0 & 73.5$_{0.4}$ &     58.5 & 67.8 & 69.9 & 70.9 & 71.3 & 70.9$_{0.3}$ \\
    & Multi & \tb{58.6} & \tb{71.7} & 72.9 & 73.3 & 73.6 & 74.2$_{0.5}$ &      \tb{58.6} & 67.9 & 69.8 & 70.6 & 70.8 & 71.1$_{0.6}$ \\
    & Imitation & 56.6 & 69.1 & 72.4 & 73.5 & 73.9 & \tb{74.6}$_{0.5}$ &      56.6 & \tb{68.7} & \tb{72.6} & \tb{74.4} & \tb{75.5} & \tb{75.4}$_{0.5}$ \\
    & Offline RL & 58.0 & 71.6 & \tb{74.0} & \tb{74.7} & \tb{75.0} & \tb{74.6}$_{0.6}$ & 58.0 & 68.4 & 71.2 & 72.2 & 72.7 & 73.3$_{0.5}$ \\
    \bottomrule
  \end{tabular}}
  \caption{\small{Online agent F1$\uparrow$ on the test set. \foneat{0-4} are from model trained with seed 1 and avg$_\text{std}$ is \foneat{4} of 5 runs.}}
  \label{tab:online_evaluation}
\end{table*}

%For comparison among interaction models, we see they all improved drastically at early turns.
%But at later turns, the performance plateaued.
%This is likely due to the fact that examples with $\geq 3$ turns are rare in offline evaluation (as in Tab.~\ref{tab:dataset}).
%Specially, the \emph{Multi} model achieved the best offline task completion rate (\foneat{all}).
%We consider this is expected since it is trained towards this metric.
%The \emph{Imitation} model consistently outperforms other models on multi-turn examples (i.e., \emph{Challenging} subset), but is sub-par in one-turn examples.
%This suggests that modeling the entire sequence of annotator selections introduces noise to model, thus sacrificing one-pass prediction.
%Therefore, it is interesting to further observe them in online environment, e.g., how it compares against models that are supervised towards overall task completion.

\subsection{Online Results} \label{sec:online_results}
Tab.~\ref{tab:online_evaluation} presents online test scores.
In general, models that are supervised by action sequences (i.e., \emph{Imitation} and \emph{Offline RL}) perform better.
Both heuristics-based and neural user models are able to guide agents towards task completion.
Comparing \emph{Single}'s \foneat{0} and \emph{Imitation}'s \foneat{4}, we see that properly using interaction boosts task completion by $18$ and $31$ on the \emph{Challenging} and \emph{All} test sets.

The average \foneat{4}`s show that heuristics-based user works better, except that the \emph{Imitation} collaborates better with the neural user.
This might be attributed to the neural user is trained to mimic human command patterns which can be ambiguous and short, while heuristics are more precise while being artificial. This also implies that a large room for further improvement to the user modeling.

Overall, we can see interactive grounding is a challenging task, even on a single screen.
%The challenge comes from both agent and user modeling.
The agent modeling involves robust multimodal understanding to self-correct.
The user modeling requires controlled language generation, which is still an open problem.
The best task completion rate on the \emph{Challenging} subset is only $\sim55\%$, suggesting a large room for future improvements.

\subsection{Agent Robustness} \label{sec:agent_robustness}
We take a deeper look at agent behavior in Tab.~\ref{tab:dup_analysis}.
We observe that agents with higher F1 tend to be more robust (lower $\Gamma$).
The best agent model (\emph{Imitation}) repeats the same mistake for only $16.8\%$ on the \emph{All} test set.
However, if we ignore those examples finished in $1$ turn i.e., $T>1$ columns, the repeating rate rises to $\sim40\%$.
The \emph{Heuristics} user, while generally improves agent F1 more than the \emph{Neural} user, has a mixed robustness impact on the \emph{Imitation} and \emph{Offline RL} agents.
On weaker agents (the first $3$ rows), the \emph{Heuristics} user leads to more salient robustness.
These observations suggest improving agent F1 has a more direct and positive impact on robustness.

\begin{table}[ht]
\small
\setlength{\tabcolsep}{5pt}
\def\arraystretch{1}
  \centering
  \scalebox{0.9}{
  \begin{tabular}{l|c|c|c|c!{\vrule width 1pt}c|c|c|c}
    \toprule
    & \multicolumn{2}{c|}{Challenging} & \multicolumn{2}{c!{\vrule width 1pt}}{All} & \multicolumn{2}{c|}{Challenging (T>1)} & \multicolumn{2}{c}{All (T>1)} \\
    \midrule
    & {Heuristics} & {Neural} & {Heuristics} & {Neural} & {Heuristics} & {Neural} & {Heuristics} & {Neural} \\
    \midrule
    Single & 44.4$_{0.9}$     & 44.9$_{1.1}$      & 25.8$_{0.4}$ & 26.9$_{0.3}$ & 60.3$_{0.9}$ & 61.0$_{1.3}$ & 59.1$_{0.9}$ & 61.7$_{0.5}$ \\
    Ins-only & 37.9$_{1.4}$   & 40.5$_{1.0}$      & 21.2$_{0.4}$ & 23.4$_{0.3}$ & 51.4$_{1.3}$ & 55.0$_{0.8}$ & 51.2$_{0.5}$ & 56.4$_{0.9}$ \\
    Multi & 38.3$_{1.3}$      & 41.3$_{1.0}$      & 21.3$_{0.3}$ & 23.8$_{0.5}$ & 51.5$_{1.8}$ & 55.6$_{1.4}$ & 51.3$_{0.8}$ & 57.9$_{1.0}$ \\
    Imitation & \tb{31.0}$_{1.2}$ & \tb{28.3}$_{1.4}$    & \tb{17.6}$_{0.3}$ & \tb{16.8}$_{0.5}$ & \tb{40.7}$_{1.7}$ & \tb{37.2}$_{1.8}$ & \tb{40.7}$_{0.5}$ & \tb{38.9}$_{1.1}$\\
    Offline RL & 36.4$_{1.1}$         & 35.5$_{0.8}$      & 19.9$_{0.4}$ & 20.5$_{0.3}$ & 48.6$_{1.0}$ & 47.4$_{1.0}$ & 48.0$_{0.7}$ & 49.5$_{0.8}$ \\
    \bottomrule
  \end{tabular}}
  \caption{\small{Agent $\Gamma\downarrow$ on the test split. Results are from $5$ random runs. Smaller $\Gamma$ means more robust.}}~\label{tab:dup_analysis}
\end{table}

\subsection{Automatic Evaluation v.s. Human Evaluation}\label{sec:auto_human_eval}
To show automatic online test is a promising surrogate for human-in-the-loop evaluation, we compare \emph{Single} with \emph{Multi}\footnote{We choose these two models as a pilot study since they perform consistently different in all our metrics.} with a group of human annotators (acting as the \textit{user}) (Tab.~\ref{tab:human_evaluation}).
The results from this human-in-the-loop evaluation are generally consistent with those from the automatic evaluation (Tab.~\ref{tab:online_evaluation}).
We should also note that such human study is not meant to reflect every minor differences in automatic evaluations.
\begin{table}[ht]
\small
\setlength{\tabcolsep}{5pt}
\def\arraystretch{1}
  \centering
  \scalebox{0.9}{
  \begin{tabular}{l|c|c|c|c|c|c}
    \toprule
    {Model} & {\foneat{0}} & {\foneat{1}} & {\foneat{2}} & {\foneat{3}} & {\foneat{4}} & $\Gamma\downarrow$ \\
    \midrule
    Single & \tb{50.0} & 56.4 & 58.2 & 58.4 & 59.4 & 42.6 \\
    Multi & 49.6 & \tb{58.4} & \tb{60.4} & \tb{62.2} & \tb{62.6} & \tb{39.4} \\
    \bottomrule
  \end{tabular}}
  \caption{\small{Human-in-the-loop evaluation on $500$ examples from the \emph{All} test set. Models are trained with seed $1$.}}~\label{tab:human_evaluation}
\end{table}

\subsection{Ablation on Heuristics} \label{sec:ablation_instruction}

To show agent improves from follow-up instructions effectively, instead of overfitting potential artifacts in the dataset,
we report our ablation studies in Tab.~\ref{tab:random_heuristics}.
Specifically, we focus on the heuristics-based user since it offers well-controlled instruction generation.
We can see that random heuristics underperform by $\sim14\%$ and repeating the initial instruction is even worse.
The $\Gamma$ scores also suggest that randomly instantiated instructions are less effective in guiding the agent.

\begin{table}[ht]
\small
\setlength{\tabcolsep}{5pt}
\def\arraystretch{1}
  \centering
  \scalebox{0.9}{
  \begin{tabular}{l|c|c|c|c|c|c|l}
    \toprule
    {\emph{Multi}} & {\foneat{0}} & {\foneat{1}} & {\foneat{2}} & {\foneat{3}} & {\foneat{4}} & {avg$_\text{std}$} & $\Gamma\downarrow$ \\
    \midrule
    Heuristics & 25.2 & \tb{47.8} & \tb{50.9} & \tb{51.7} & \tb{52.4} & - & \tb{40.0} \\
    Random$_\text{(5 runs)}$ & 25.2 & 32.7 & 34.3 & 34.7 & 35.1 & 35.6$_{0.9}$ & 51.6$_{1.5}$ \\
    Repeat $c_0$ & 25.2 & 29.3 & 30.9 & 31.6 & 32.0 & - & - \\
    \bottomrule
  \end{tabular}}
  \caption{\small{Ablation of instructions using heuristics-based user model for the \emph{Multi} agent on the \emph{Challenging} test set.
  The \emph{Multi} is trained with seed $1$.
  \emph{Random}: randomly instantiated heuristics for $c_{t>0}$ across $5$ seeds.}}~\label{tab:random_heuristics}
\end{table}

\subsection{Error Analysis} \label{sec:error_analysis}
We manually analyze errors from the best agent (\emph{Imitation}).
In Tab.~\ref{tab:error_analysis}, we inspect $30$ failed development examples (i.e., unfinished after $5$ turns) that are subject to the \emph{Neural} user.
Due to the role interplay, we also count problematic commands.
%For the agent, we look at the type of understanding required with respect to the instruction for each turn.
%For the user, we check whether the instruction is informative in the context.
We observe that the user model sometimes issues repetitive or uninformative instructions starting from the $3$rd turn, leading the agent to the same wrong selection. This might be caused by the data sparsity for examples with $\geq3$ turns.
%, i.e., less than 3\% of the data has more than two turns.
%Such cases happen in average of $2.4$ turns for each failed example.

\begin{table}[h]
\small
\setlength{\tabcolsep}{3pt}
\def\arraystretch{1}
  \centering
  \scalebox{0.9}{
  \begin{tabular}{l|c|c|c|c!{\vrule width 1pt}c|c}
    \toprule
    & \multicolumn{4}{c!{\vrule width 1pt}}{Agent} & \multicolumn{2}{c}{User} \\
    \midrule
    Incapabilities & {text} & {icon} & {UI layout} & {pos/dir} & {wrong $c_t$} & {stale $c_t$} \\
    \midrule
    \#Example & 6 & 7 & 9 & 7 & 15 & 27 \\
    \bottomrule
  \end{tabular}}
  \caption{\small{Major error categories of the \emph{Imitation} model on $30$ failed development examples ($150$ turns).
  \emph{stale $c_t$}: repetitive/uninformative instruction.
  Model is trained with random seed $1$.}}~\label{tab:error_analysis}
\end{table}

%% file: conclusion.tex
\section{Conclusions}
In this paper, we presented \myname, a novel and challenging task for multimodal grounding on UI.
\myname requires a grounding agent being able to correct its own prediction,
and allows a user to guide the agent via natural language instructions.
We contribute a new dataset for the task, investigate various modeling options for the agent and developed multiple evaluation strategies including two user models so as to enable automatic online testing.
We found that interaction greatly improves grounding accuracy in the UI domain.
Our experiments and analyses also suggest large room for grounding performances, even on a seemingly easy single screen task, which calls for future investigation.
Our work contributes to the general effort of multimodal language understanding and its robustness by enabling synchronized multi-turn interactivity.

%% file: appendix.tex
\section{Labeling Interface} \label{sec:labeling_interface}
\begin{figure}[ht]
 \centering
 \vspace*{5pt}%
 \hspace*{\fill}% 
  \begin{subfigure}[b]{0.8\textwidth}
  \centering
    \includegraphics[height=0.35\textheight]{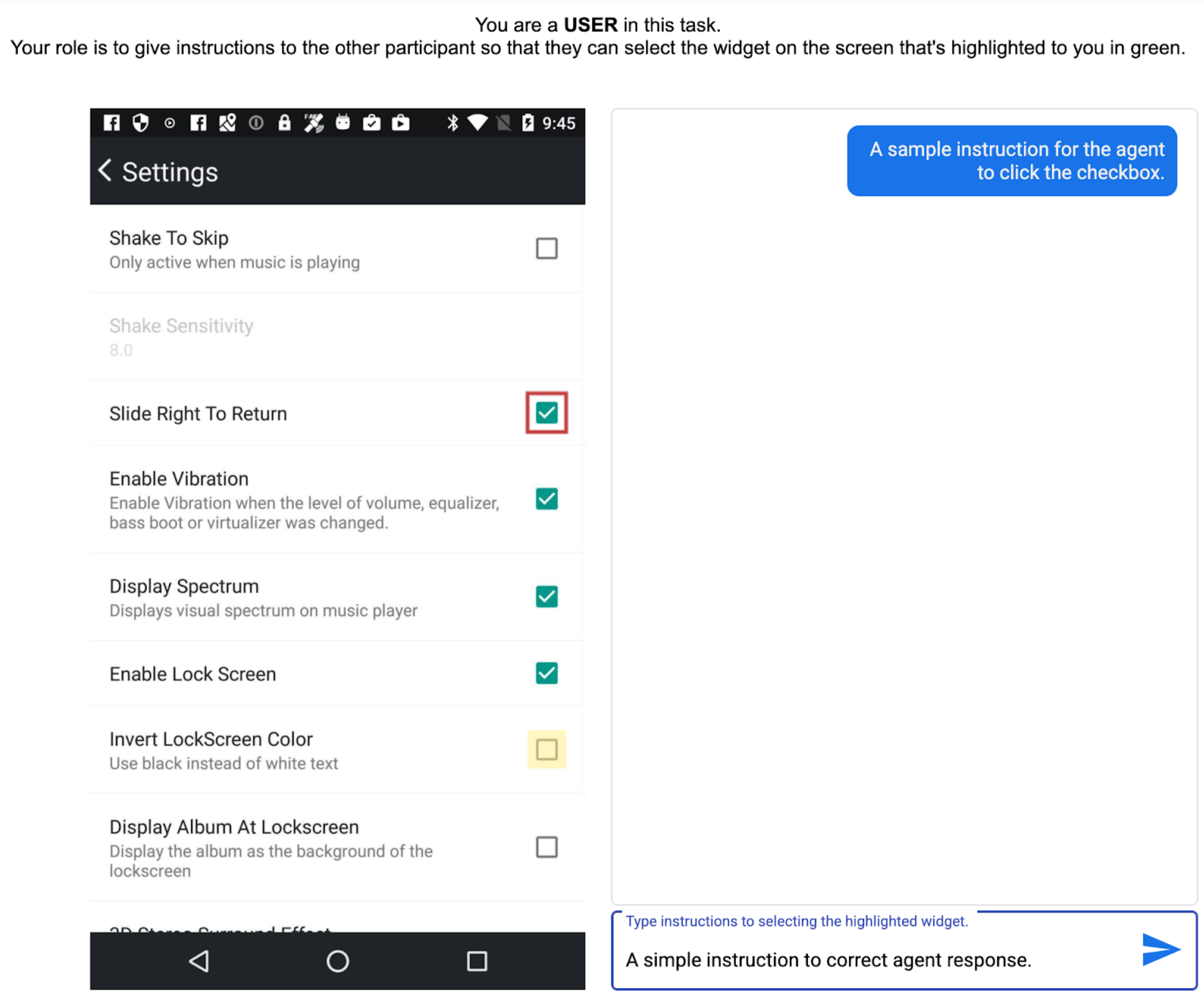}
    \caption{The user sees the target object (boxed in red) and the agent selection in the previous round (boxed in yellow). The user can issue commands in the message box.}
  \end{subfigure}%
  \hspace*{\fill}%          % empty line absolutely necessary!

  \vspace*{2pt}%  

  \hspace*{\fill}%  
   \begin{subfigure}[b]{0.8\textwidth}
   \centering
    \includegraphics[height=0.35\textheight]{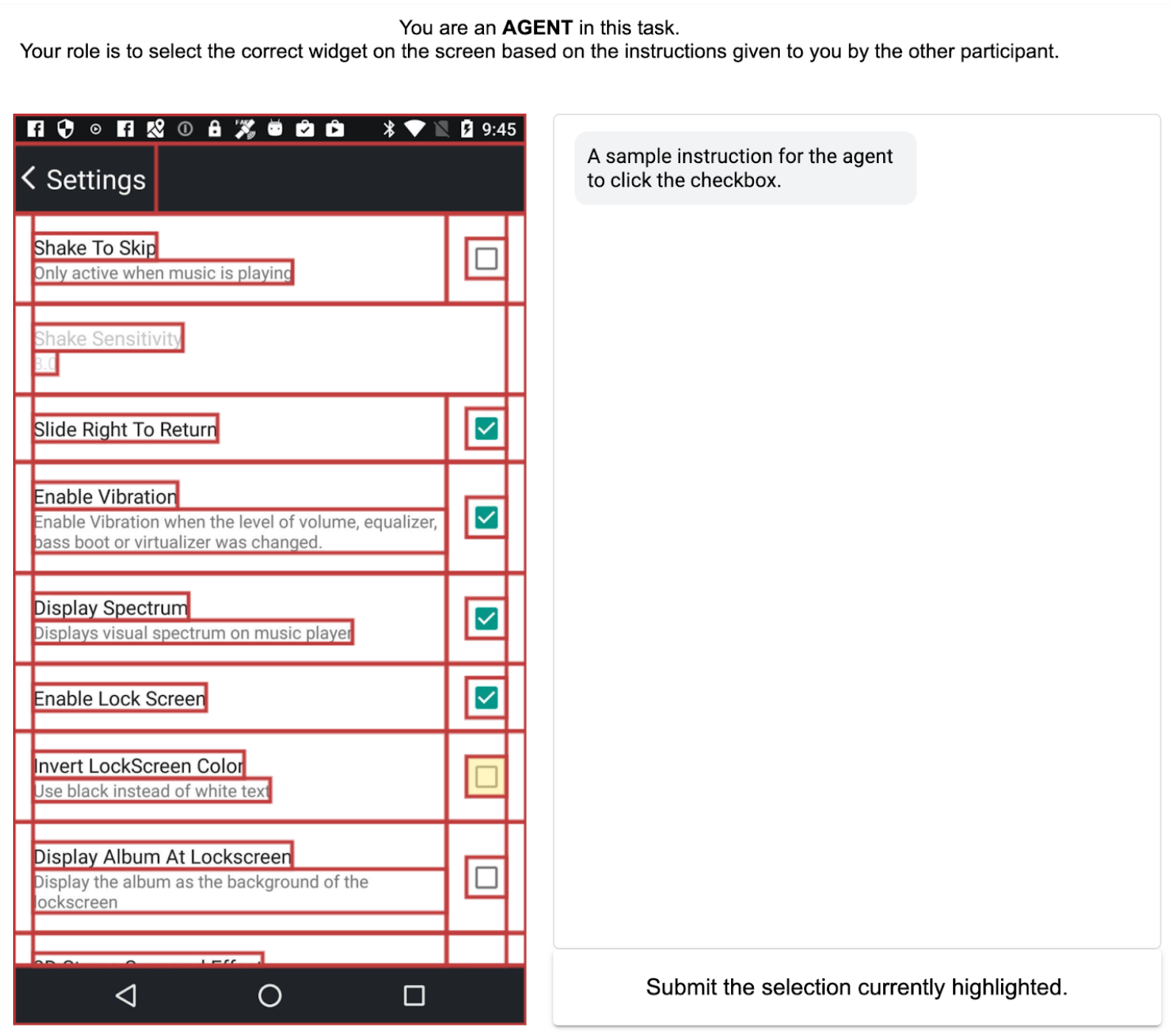}
    \caption{The agent sees the user commands, and all the available candidates (clickable objects) on the screen, which are all boxed in red, and the current selection boxed in yellow.}
  \end{subfigure}%
  \hspace*{\fill}%          % empty line absolutely necessary!
  %\captionsetup{skip=4pt}% % required to hide the figure (main) lable!
  \caption{\small{\myname annotation interfaces consist of a user view and an agent view.}}~\label{fig:annotation_interface}
\end{figure}

Fig.~\ref{fig:annotation_interface} presents the user and agent views in our data collection interface. In the user view, the user can send commands in the message box, to instruct the agent to select the target object as highlighted by a red bounding box on the UI screen. On the agent's view, the agent annotator can respond the user request by performing object selection on the UI screen, which has all the clickble objects highlighted. But there is no indication of the target object so the agent annotator has to guess from the user instruction. The agent is not allowed to text back to the user. The agent's current selection is reflected on the UI screen so the user understands how to further instruct the agent. The annotation task is designed based on the eyes-on hands-free situation of mobile interaction.

\section{Details of the Labeling Task}
The labelers of the task were native English speakers and had experience using mobile phones. They were trained with a few pilot tasks to get familiar with the task, during which we also improved the labeling interface and the guidelines based on labelers' feedback. The dataset was completed by $30$ labelers in $10$ batches. The labeling quality was monitored by sampling examples from each batch for manual examination.

\section{View Hierarchy Features}\label{sec:vh_features}
Tab.~\ref{tab:feat_visual_structure} lists the complete features we used.
We unify each feature into a real-valued vector.
For text attributes (e.g., \emph{text}), we max-pool their non-contextualized token embeddings, which are randomly initialized and trained.
For discrete-valued attributes (e.g., \emph{type}), we use a trainable vector for each possible value.
The ordering of objects in Eq.~\ref{eq:encoder}\&\ref{eq:grounding_decoder} follows the pre-order traversal in the view hierarchy tree.

\begin{table}[h]
\centering
\small
\def\arraystretch{0.9}
\begin{tabular}{l|l}
    \toprule
    Feature & Example \\
    \midrule
    bounding box & [xmin, xmax, ymin, ymax] \\
    leaf & true/false \\
    type & button/checkbox/... \\
    clickable & true/false \\
    text & email address/passcode \\
    resource id & login\_icon \\
    dom & [pre/post-order index] \\
    \bottomrule
\end{tabular}
\caption{\small{Features $\psi$ used for visual structure.}}
\label{tab:feat_visual_structure}
\end{table}

\section{Hyperparameters \& Training}\label{sec:hyperparameters}
For all our agent models, we use the same configurations, which are grid-searched based on models' offline validation performances. Our hyperparameters are chosen from the best offline development F1 scores. For the number of self-attention modules, we grid-searched in $\{1,2,4,6\}$, which resulted in $2$ hidden layers for the user interface Transformer encoder and $6$ hidden layers for the grounding decoder.
Each self-attention module uses 8-head self and encoder-decoder attention with a $256$ hidden size.
The dropout rate for attention and MLP layers is $0.1$, which is grid-searched in $\{0.1, 0.2, 0.5\}$.
For learning rate, we grid-searched from $\{1e\texttt{-}3, 3e\texttt{-}4, 1e\texttt{-}4, 3e\texttt{-}5, 1e\texttt{-}5\}$, and use $3e\texttt{-}4$ with linear warmup with cosine annealing for the first $10$k steps. All the models are trained to $100$k steps with a batch size of $128$ on a $32$-core Google Cloud TPUv3.
Models are evaluated every $1$k steps and the version with the best development score is saved.
The training time is around $8$ hours.

Our neural user model has the same grid-searched configuration as the agent, i.e., $2$ encoder layers, $6$ decoder layers, $0.1$ for dropout, and the same warmup scheduling.
The best learning rate is $1e\texttt{-}4$.
Different from the agent model, we found the neural user model's development CIDEr score quickly drops after $6$k steps, possibly due to overfitting and data sparsity, thus its training early-stops there.

% \begin{table}[h]
% \centering
% \begin{tabular}{l|l}
%     \toprule
%     Hyperparameter & Value \\
%     \midrule
%     Learning rate & 3e-4 \\
%     Dropout & 0.1 \\
%     Encoder layers & 2 \\
%     Decoder layers & 6 \\
%     Warm-up & linear*cosine \\
%     Batch size & 128 \\
%     Training steps & 100k with early stopping \\
%     Training time & 1 day \\
%     Accelerator type & 32-core TPU \\
%     \bottomrule
% \end{tabular}
% \caption{Model hyperparameters.}
% \label{tab:hyperparameters}
% \end{table}

\section{Examples in the \myname Dataset} \label{sec:mug_examples}
\begin{figure}[ht]
 \centering
 \vspace*{5pt}%
 \hspace*{\fill}% 
  \begin{subfigure}{\textwidth}
    \includegraphics[height=0.22\textheight]{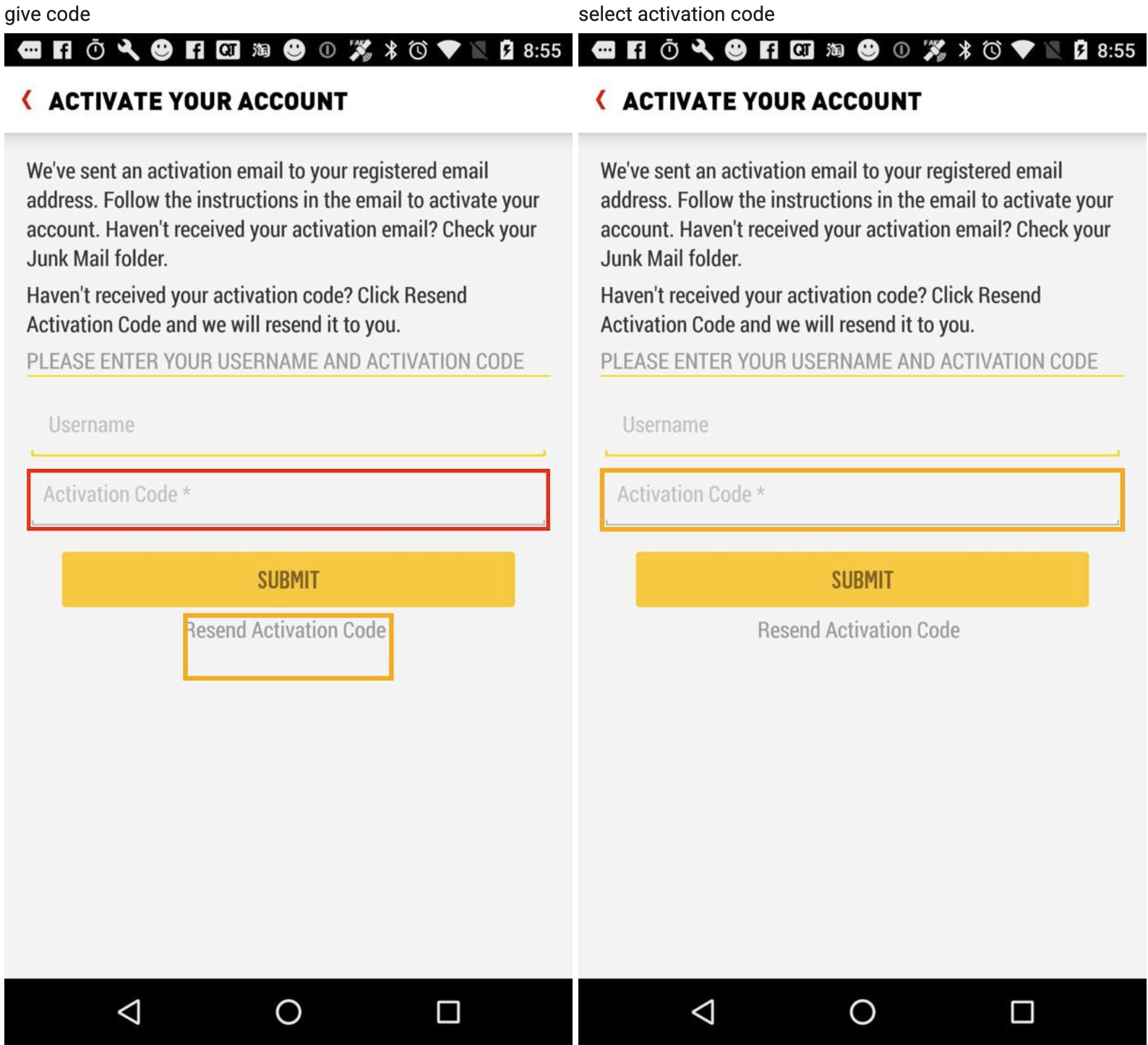}
    \label{fig:user_view}
  \end{subfigure}%
  \hspace*{\fill}%          % empty line absolutely necessary!

  \vspace*{2pt}%  

  \hspace*{\fill}%  
   \begin{subfigure}{\textwidth}
    \includegraphics[height=0.213\textheight]{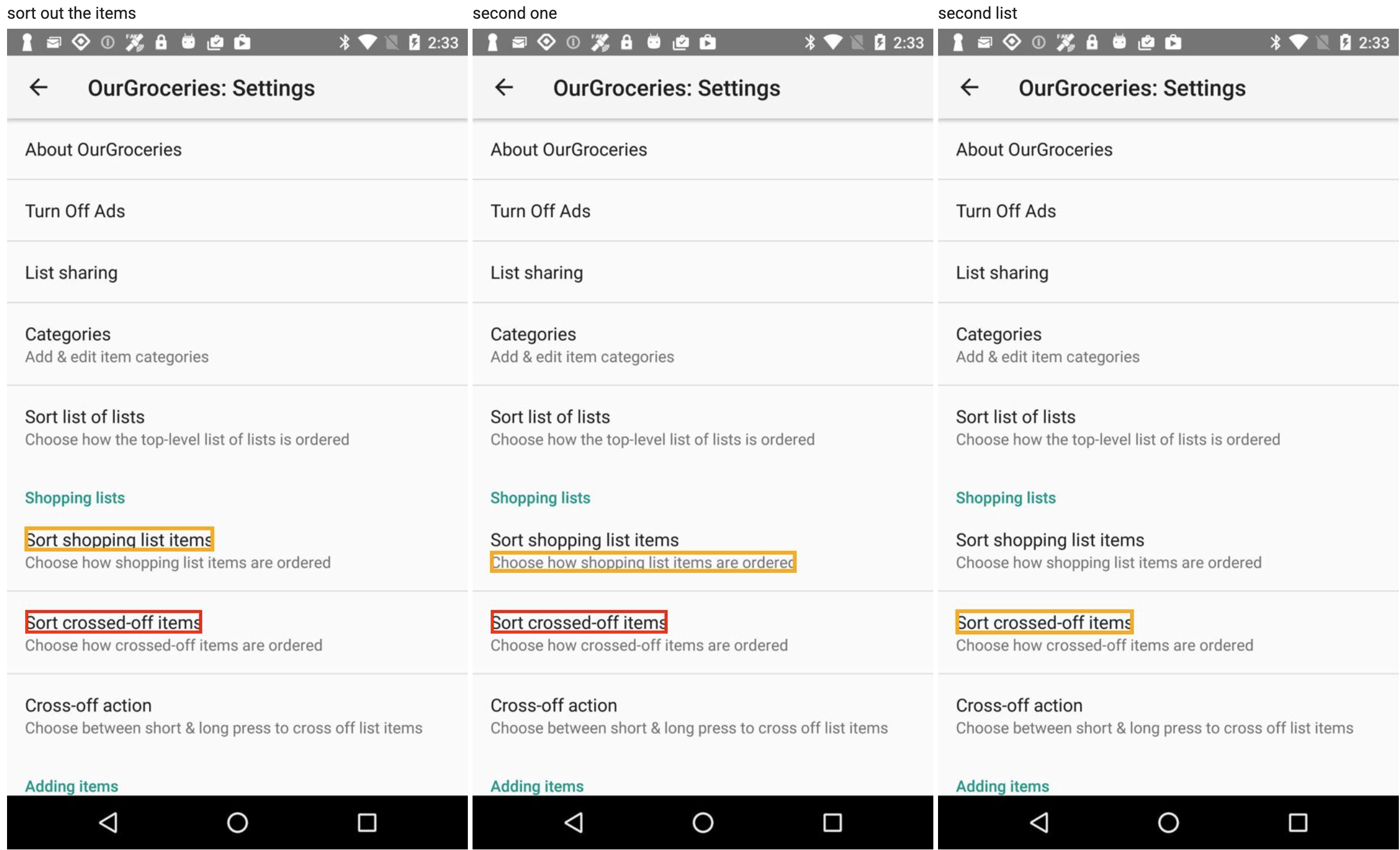}
    \label{fig:agent_view}
  \end{subfigure}%
  \hspace*{\fill}%          % empty line absolutely necessary!

  \vspace*{2pt}%

  \hspace*{\fill}%  
  \begin{subfigure}{\textwidth}
    \includegraphics[height=0.22\textheight]{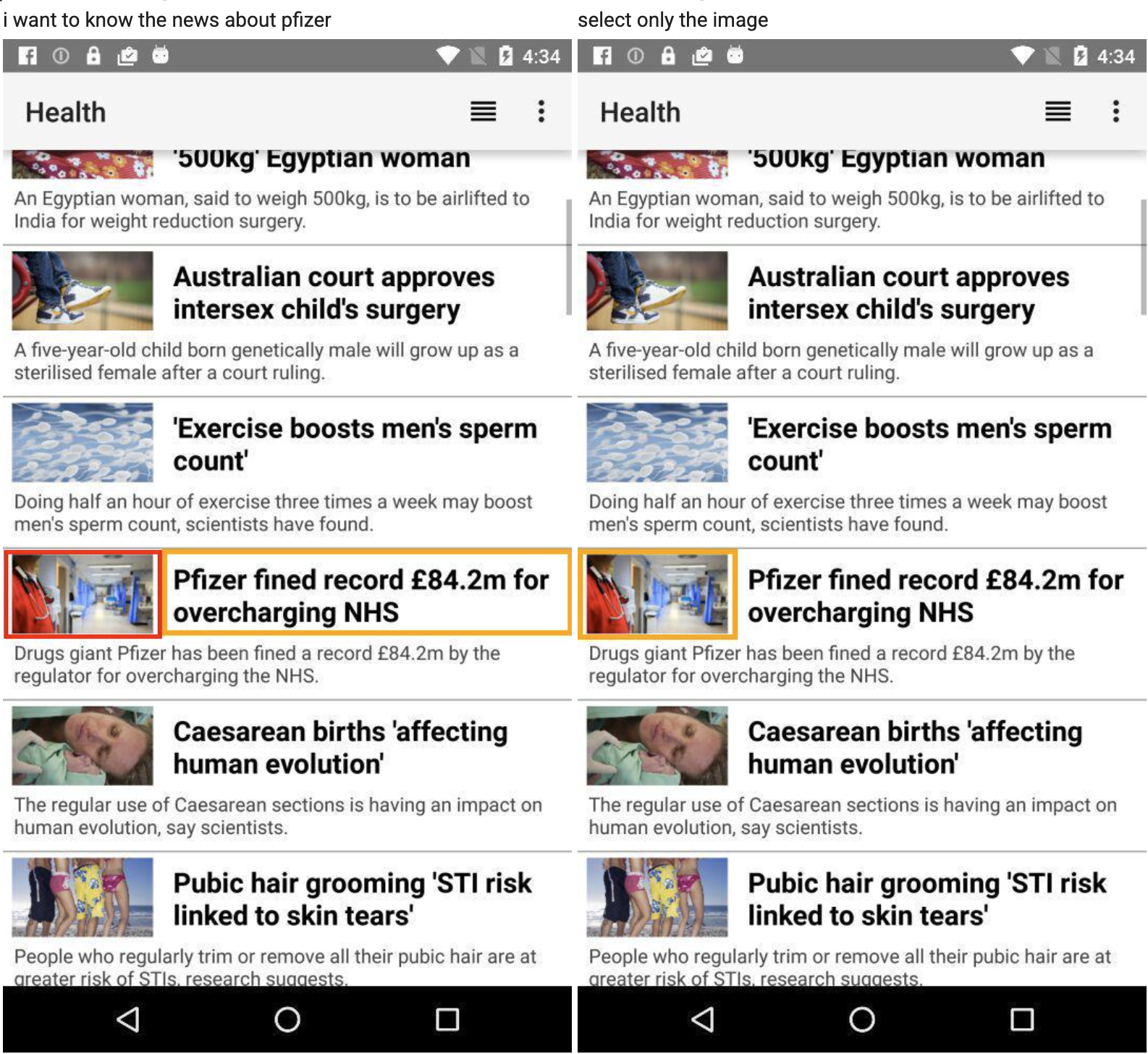}
    \label{fig:agent_view}
  \end{subfigure}%
  \hspace*{\fill}%
  
  \vspace*{2pt}%

  \hspace*{\fill}%  
  \begin{subfigure}{\textwidth}
    \includegraphics[height=0.22\textheight]{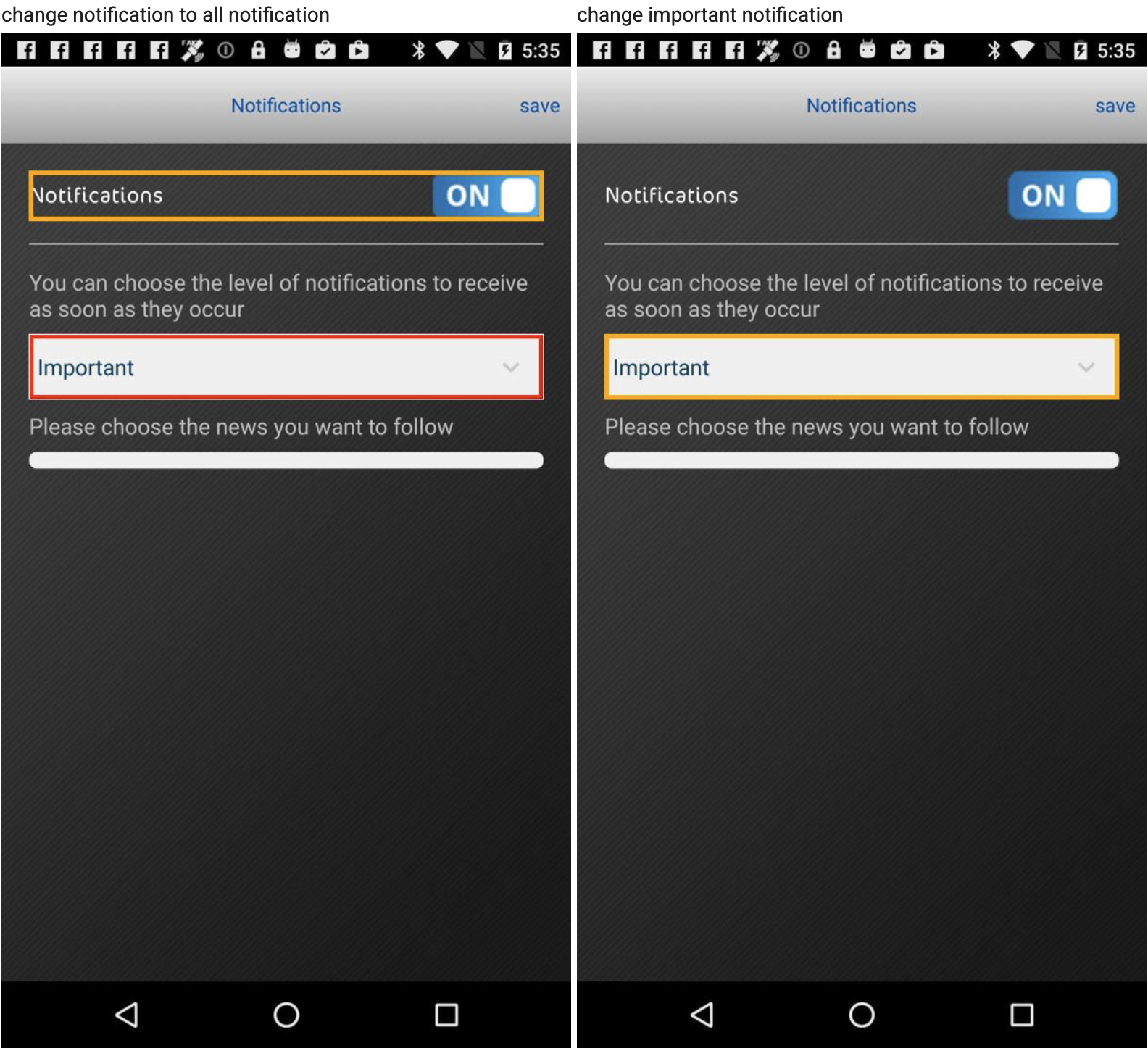}
    \label{fig:agent_view}
  \end{subfigure}%
  \hspace*{\fill}%
  
%  \captionsetup{skip=4pt, format=nocap}% % required to hide the figure (main) lable!
  \caption{\small{\myname examples 1-4. Instructions are at top of each turn. Agent selection is in \protect\inlinegraphics{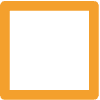} and target is in \protect\inlinegraphics{figures/icon_target.png}.}}
  \label{fig:mug_ex1to4}
\end{figure}

\begin{figure}[ht]
 \centering
 \vspace*{5pt}%
 \hspace*{\fill}% 
  \begin{subfigure}{\textwidth}
    \includegraphics[height=0.22\textheight]{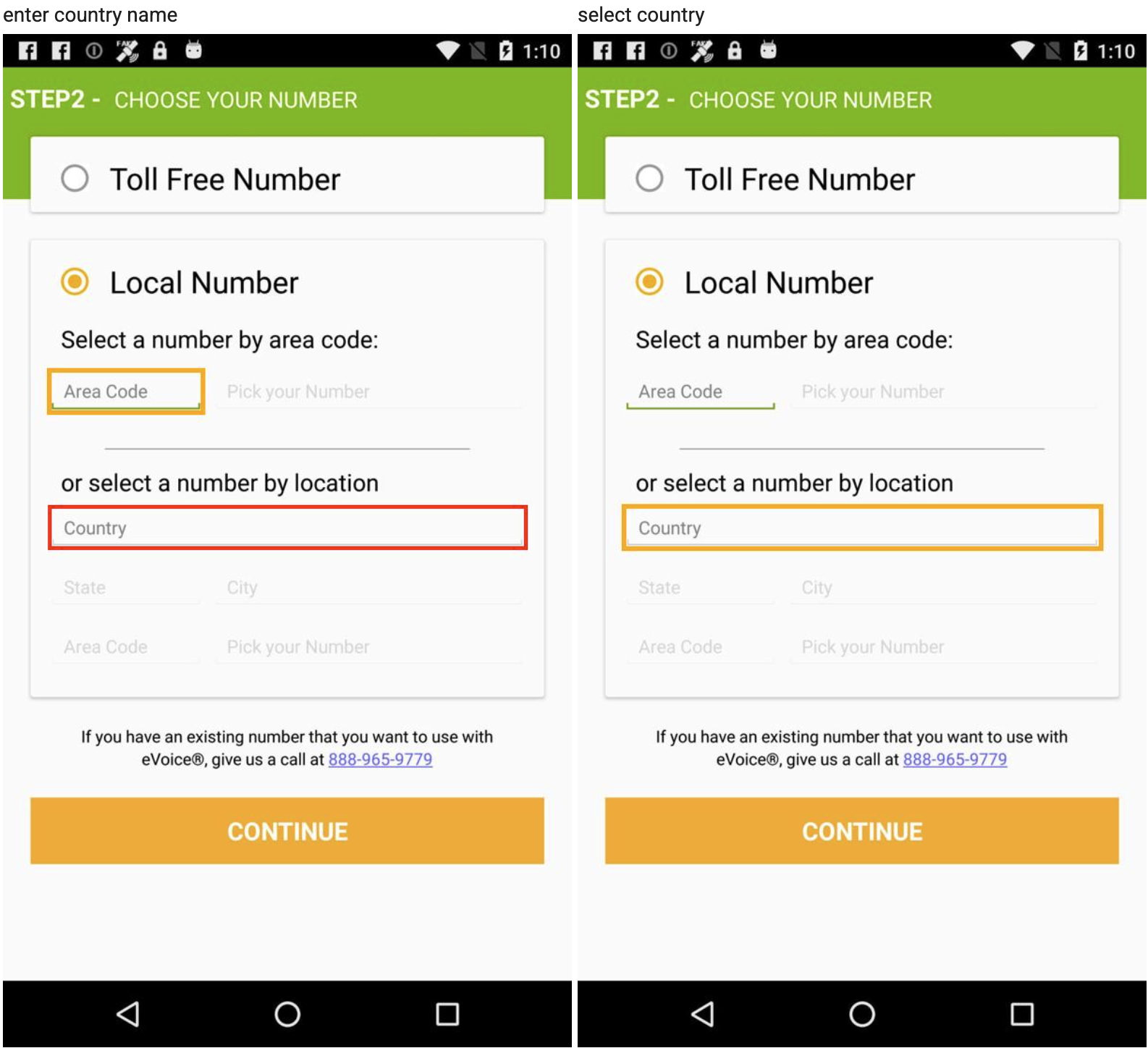}
    \label{fig:user_view}
  \end{subfigure}%
  \hspace*{\fill}%          % empty line absolutely necessary!

  \vspace*{2pt}%  

  \hspace*{\fill}%  
   \begin{subfigure}{\textwidth}
    \includegraphics[height=0.213\textheight]{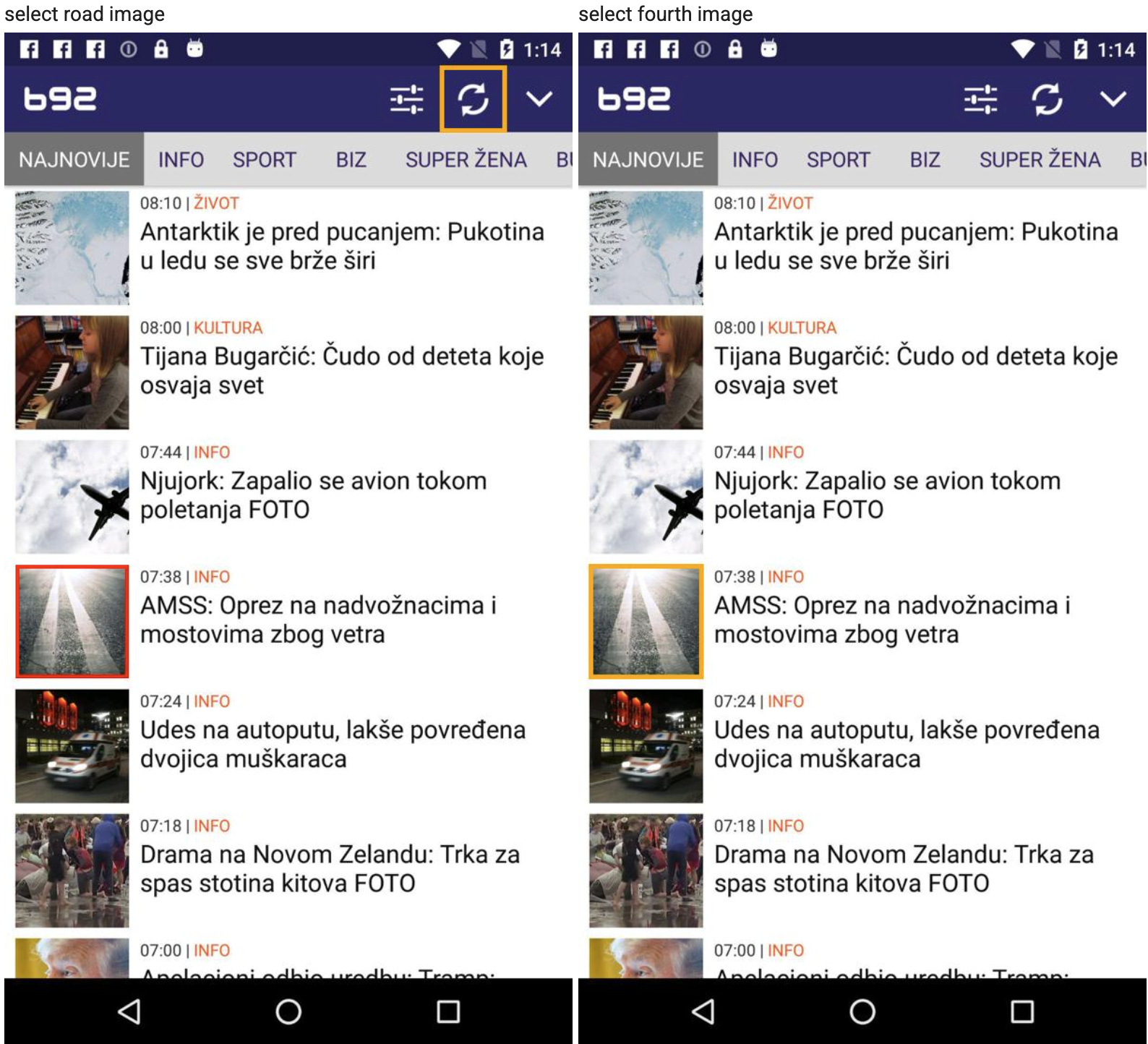}
    \label{fig:agent_view}
  \end{subfigure}%
  \hspace*{\fill}%          % empty line absolutely necessary!

  \vspace*{2pt}%

  \hspace*{\fill}%  
  \begin{subfigure}{\textwidth}
    \includegraphics[height=0.22\textheight]{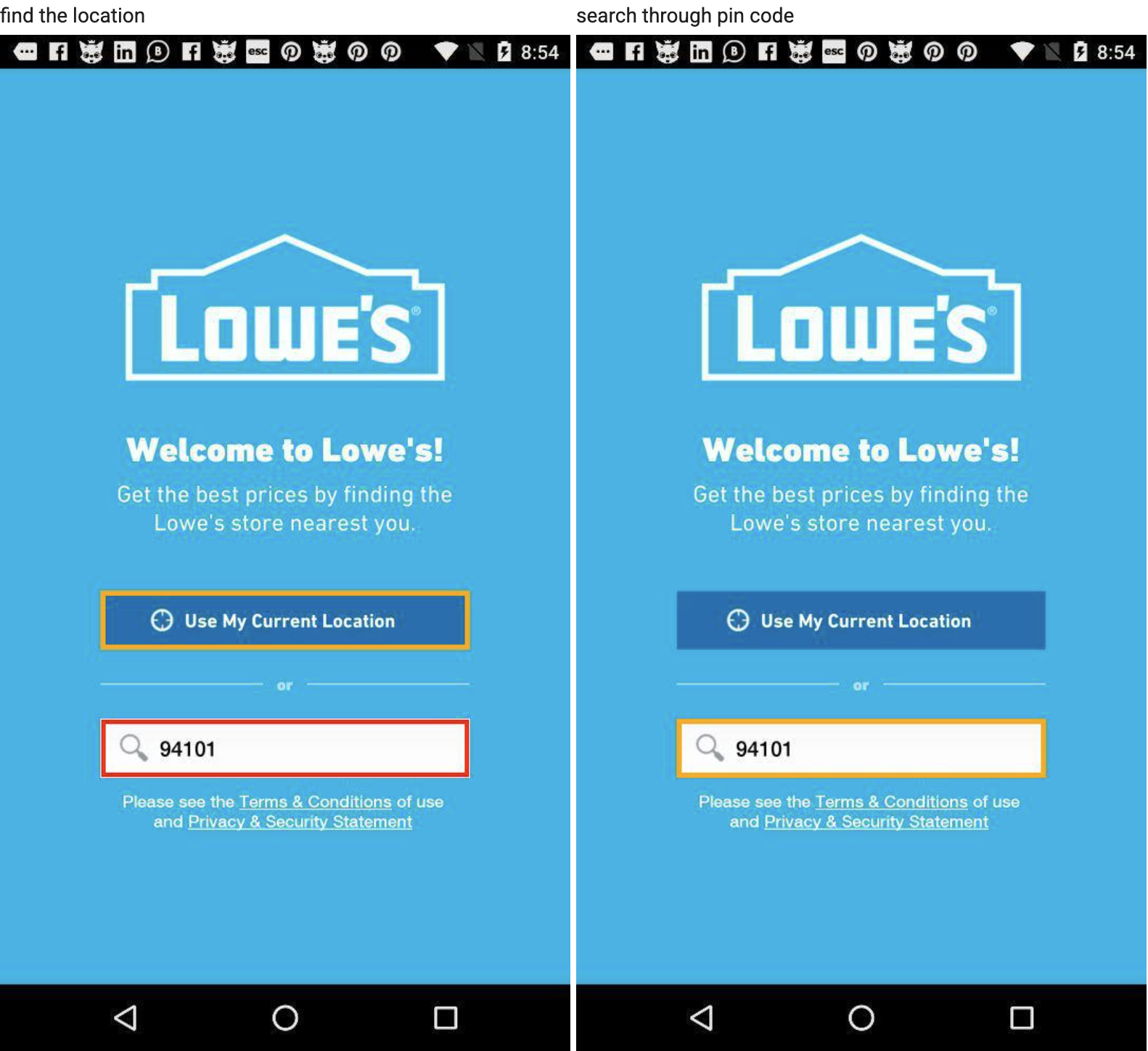}
    \label{fig:agent_view}
  \end{subfigure}%
  \hspace*{\fill}%
  
  \vspace*{2pt}%

  \hspace*{\fill}%  
  \begin{subfigure}{\textwidth}
    \includegraphics[height=0.22\textheight]{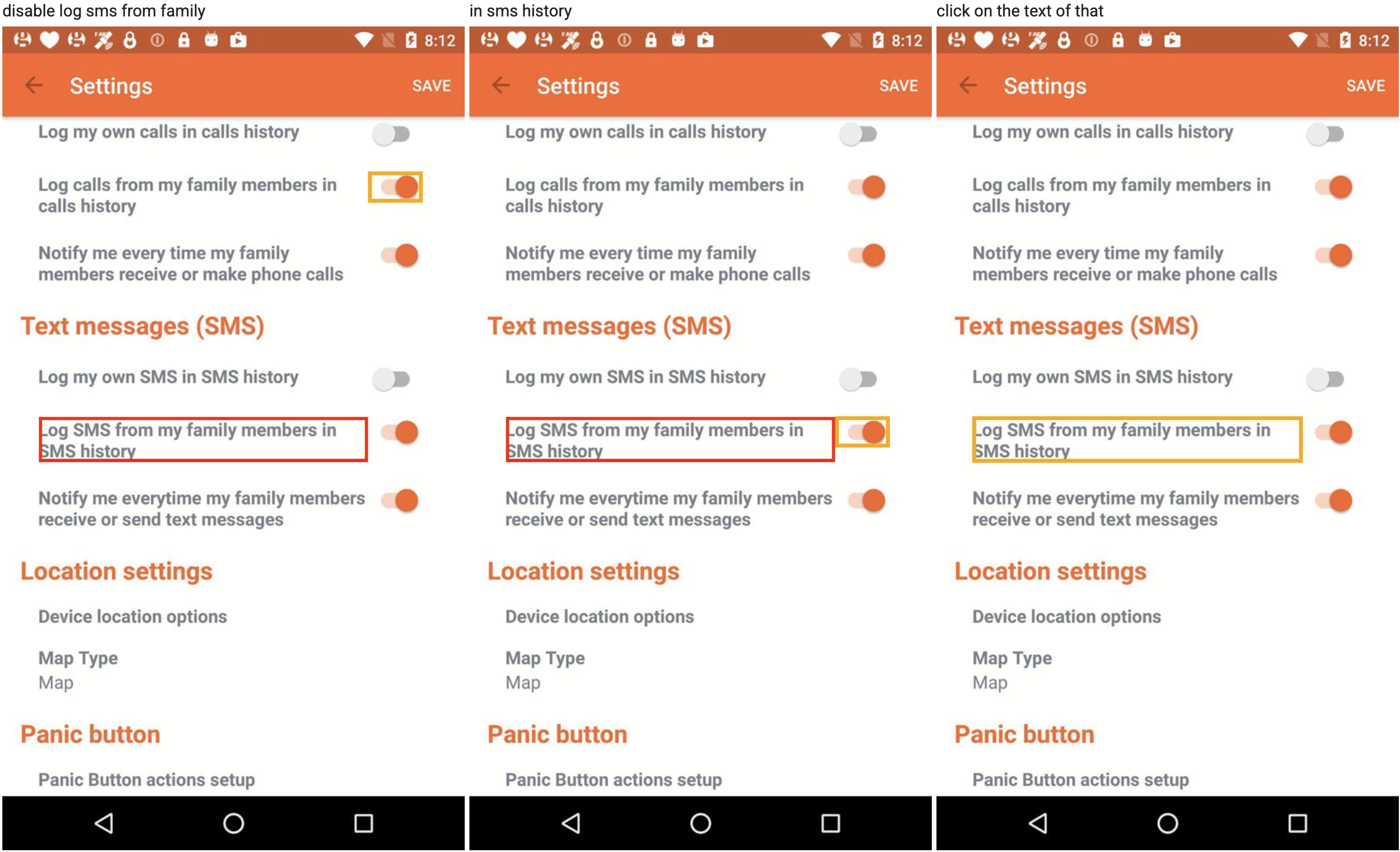}
    \label{fig:agent_view}
  \end{subfigure}%
  \hspace*{\fill}%
  
%  \captionsetup{skip=4pt, format=nocap}% % required to hide the figure (main) lable!
  \caption{\small{\myname examples 5-8. Instructions are at top of each turn. Agent selection is in \protect\inlinegraphics{figures/icon_orange.png} and target is in \protect\inlinegraphics{figures/icon_target.png}.}}~\label{fig:mug_ex5to8}
\end{figure}

We present some examples from the \myname dataset in Fig.~\ref{fig:mug_ex1to4} and \ref{fig:mug_ex5to8}.
Each example contains instructions and selections from human user and agent annotators.

\section{Prediction Examples} \label{sec:pred_examples}
\begin{figure}[ht]
 \centering
 \vspace*{5pt}%
 \hspace*{\fill}% 
  \begin{subfigure}{\textwidth}
    \includegraphics[height=0.22\textheight]{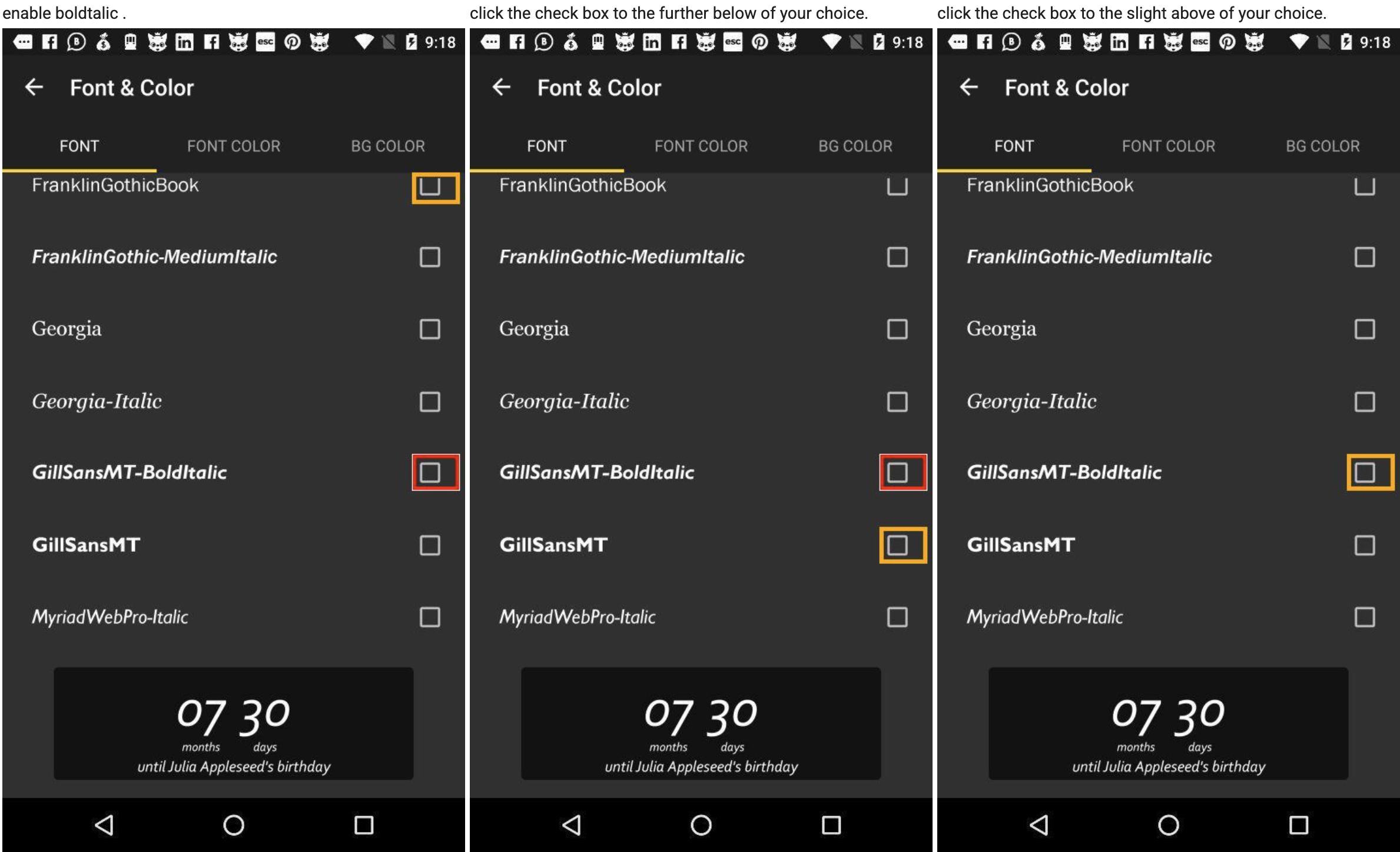}
    \label{fig:user_view}
  \end{subfigure}%
  \hspace*{\fill}%          % empty line absolutely necessary!

  \vspace*{2pt}%  

  \hspace*{\fill}%  
   \begin{subfigure}{\textwidth}
    \includegraphics[height=0.22\textheight]{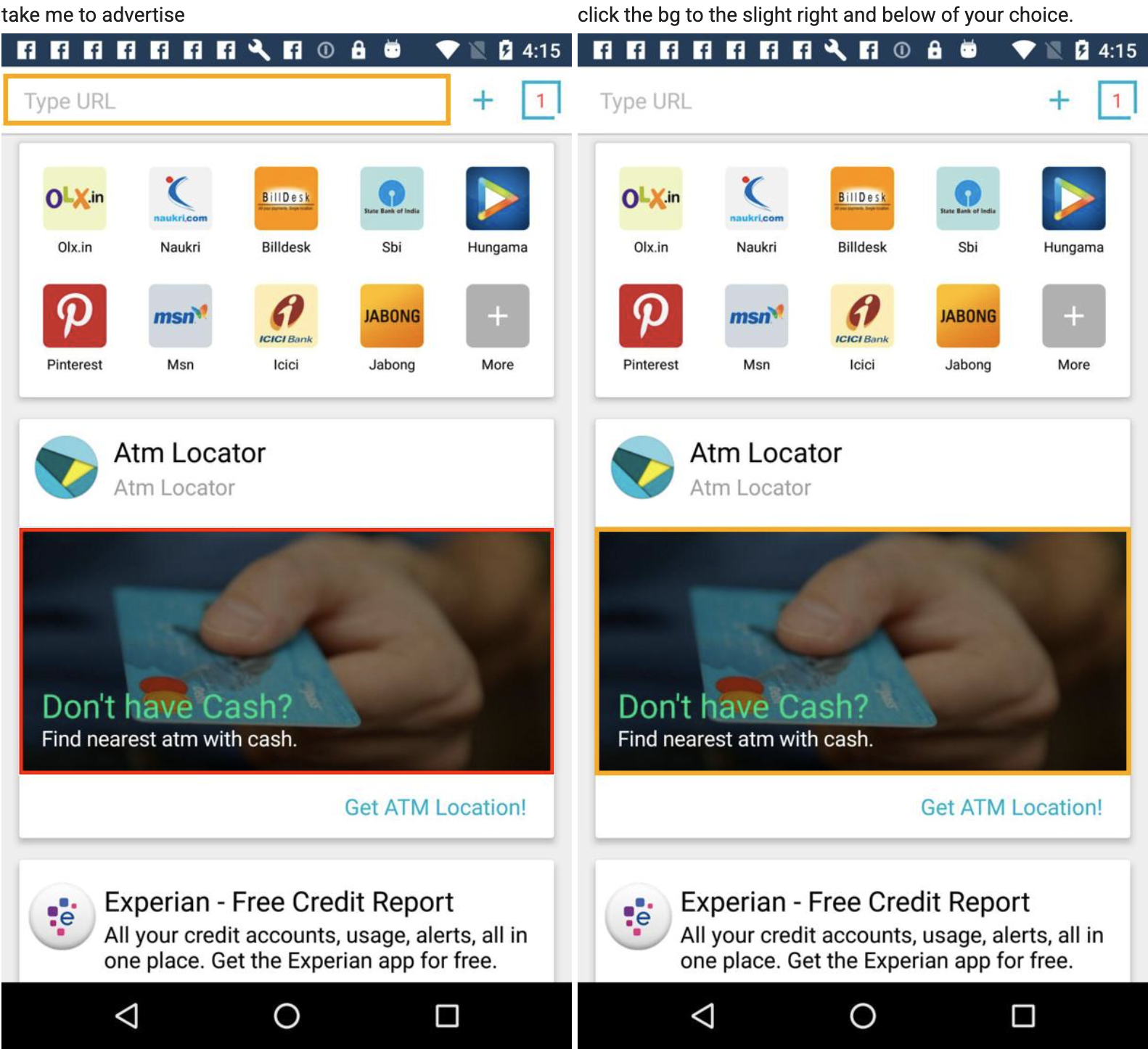}
    \label{fig:agent_view}
  \end{subfigure}%
  \hspace*{\fill}%          % empty line absolutely necessary!

  \vspace*{2pt}%

  \hspace*{\fill}%  
  \begin{subfigure}{\textwidth}
    \includegraphics[height=0.22\textheight]{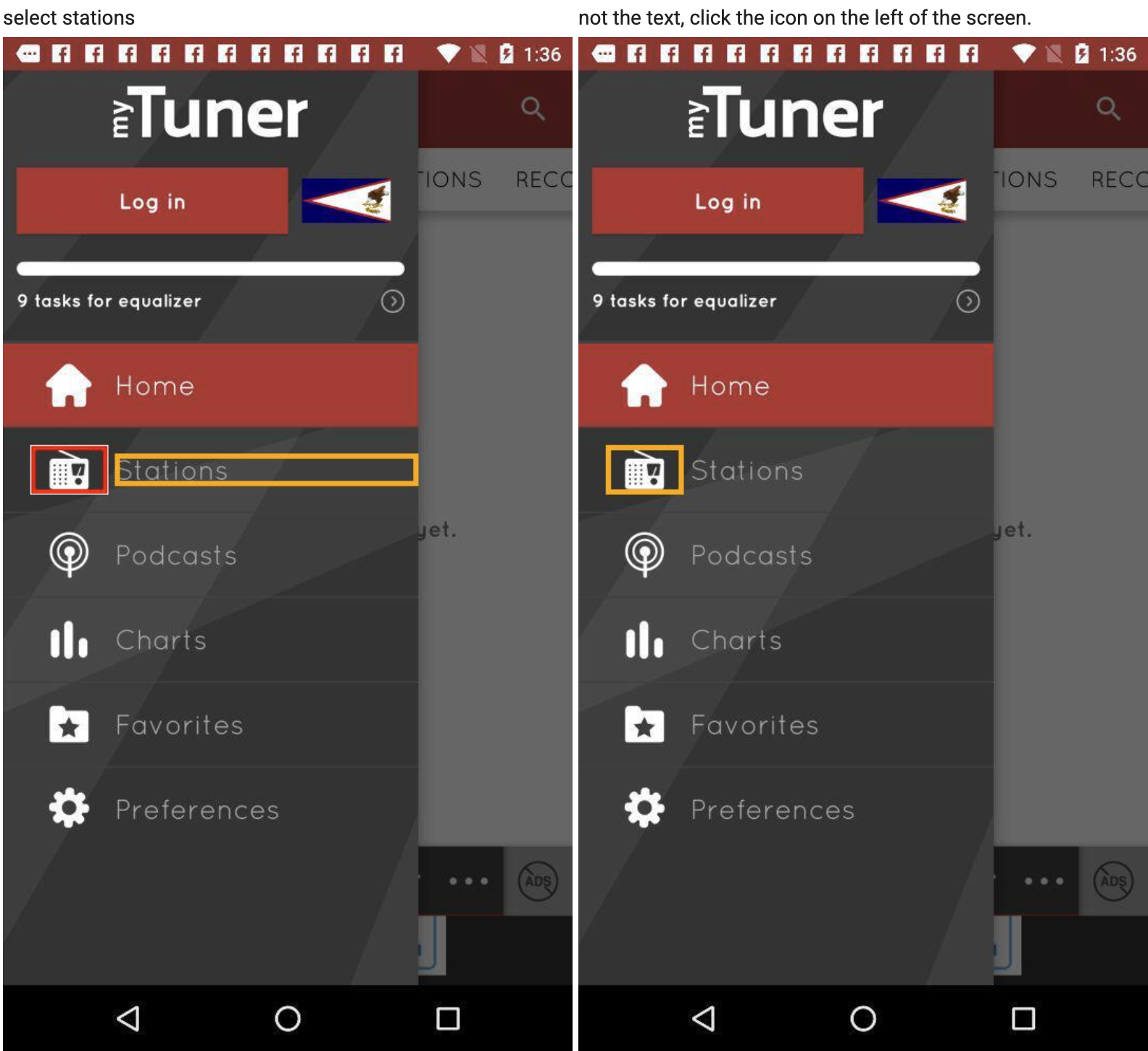}
    \label{fig:agent_view}
  \end{subfigure}%
  \hspace*{\fill}%
  
  \vspace*{2pt}%

  \hspace*{\fill}%  
  \begin{subfigure}{\textwidth}
    \includegraphics[height=0.22\textheight]{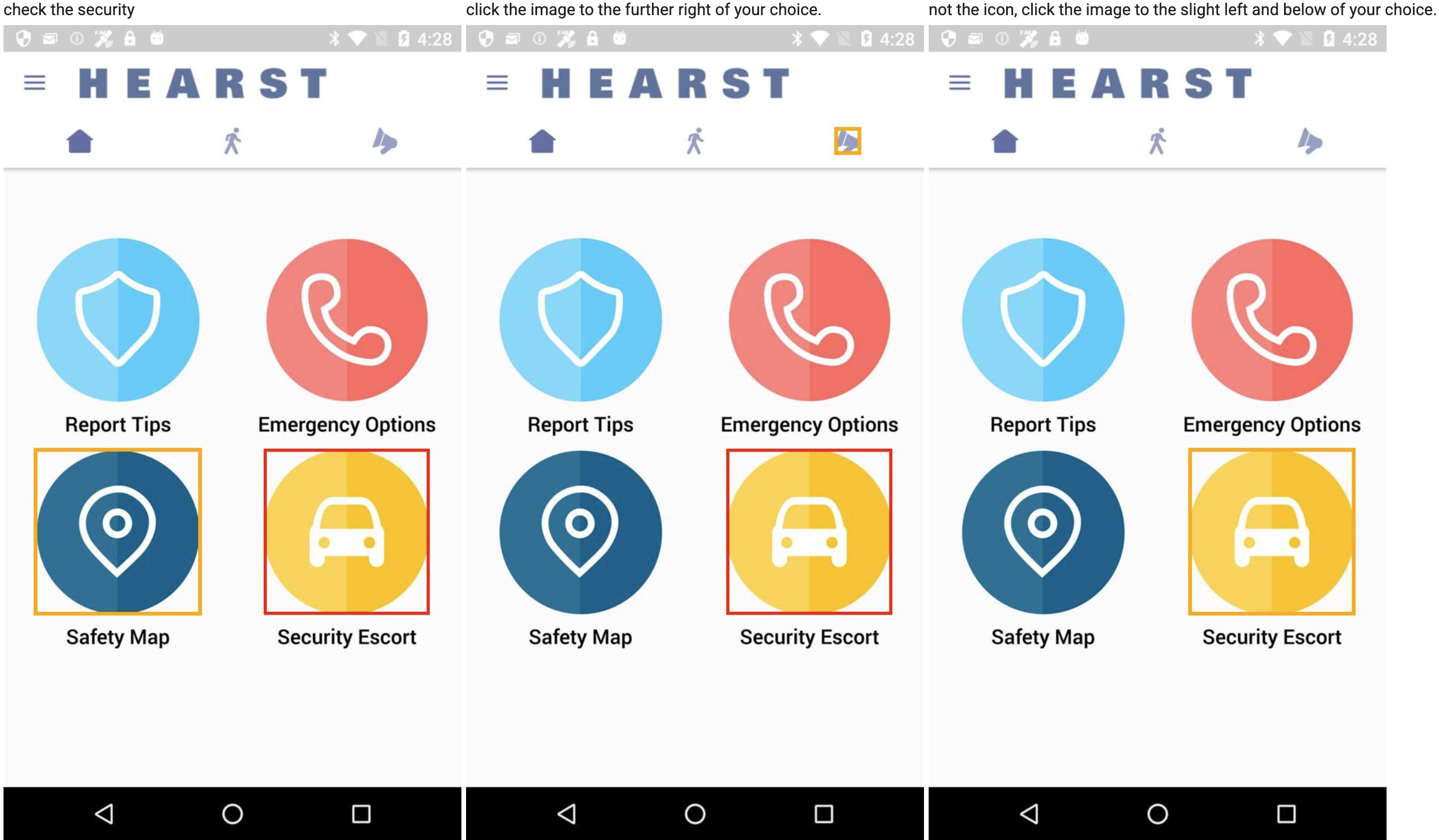}
    \label{fig:agent_view}
  \end{subfigure}%
  \hspace*{\fill}%
  
%  \captionsetup{skip=4pt, format=nocap}% % required to hide the figure (main) lable!
  \caption{\small{\tb{Completed} examples by the \emph{Imitation} agent following the instructions generated by the \tb{Heuristic} user.}}~\label{fig:completed_heu}
\end{figure}

\begin{figure}[ht]
 \centering
 \vspace*{5pt}%
 \hspace*{\fill}% 
  \begin{subfigure}{\textwidth}
    \includegraphics[height=0.22\textheight]{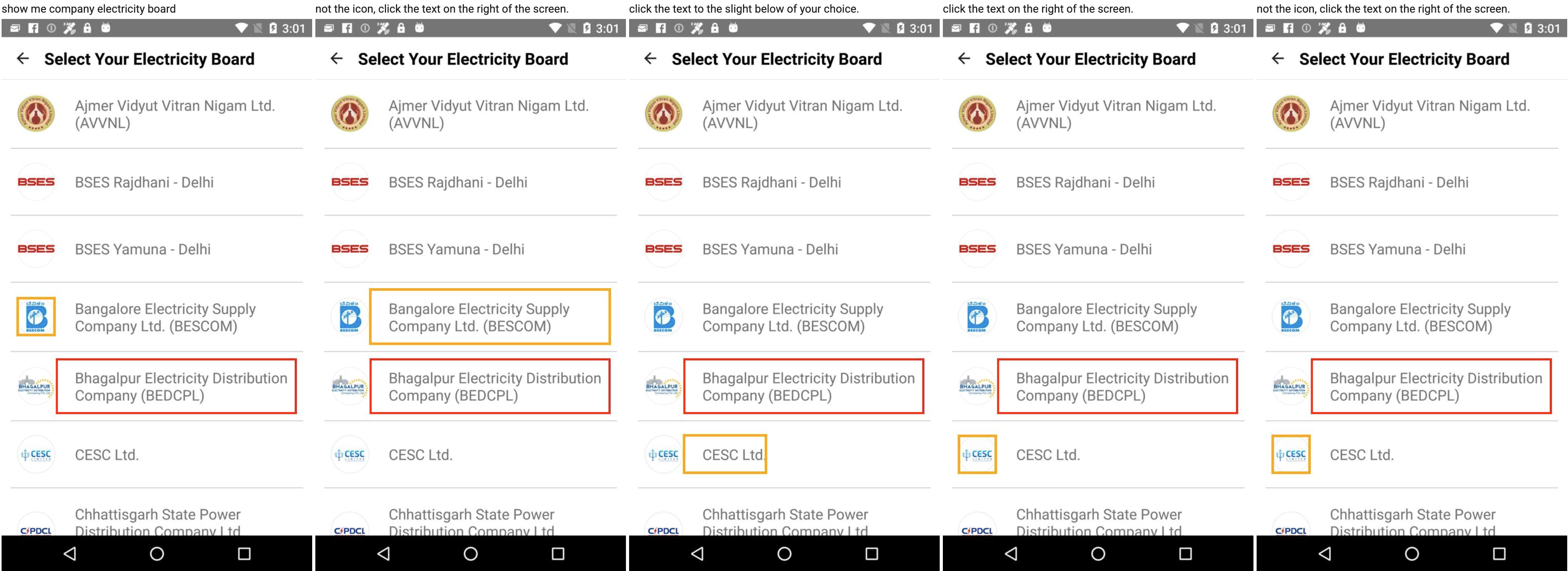}
    \label{fig:user_view}
  \end{subfigure}%
  \hspace*{\fill}%          % empty line absolutely necessary!

  \vspace*{2pt}%  

  \hspace*{\fill}%  
   \begin{subfigure}{\textwidth}
    \includegraphics[height=0.213\textheight]{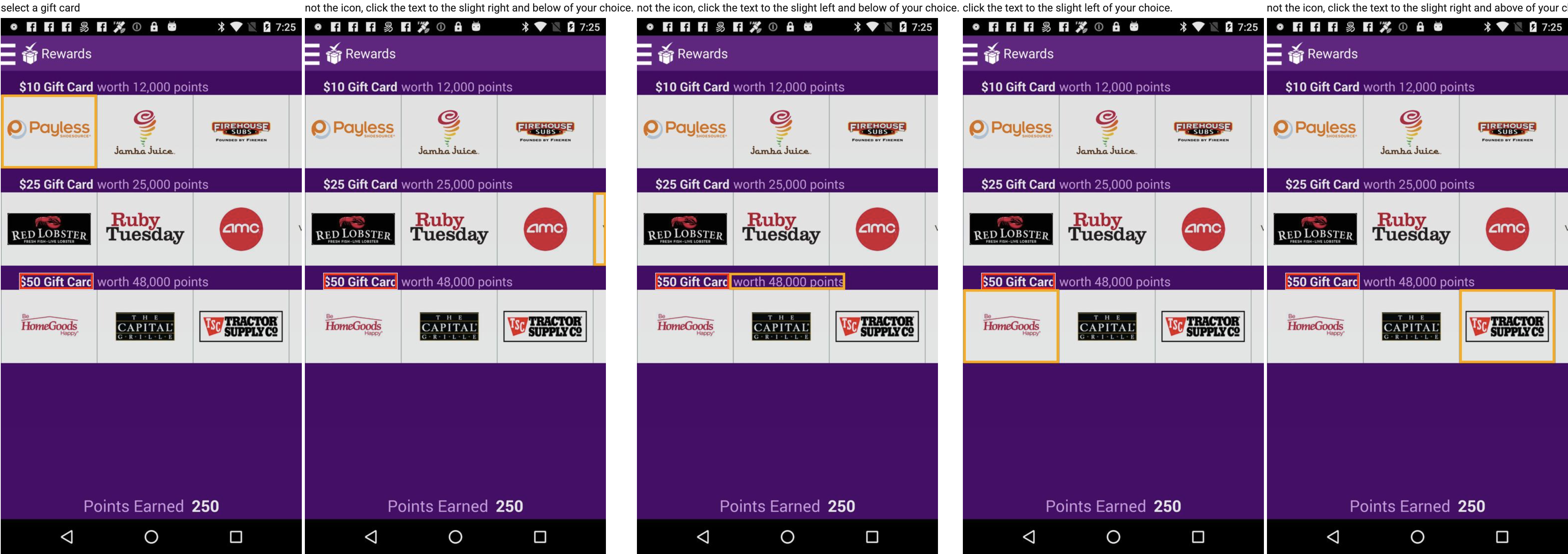}
    \label{fig:agent_view}
  \end{subfigure}%
  \hspace*{\fill}%          % empty line absolutely necessary!

  \vspace*{2pt}%

  \hspace*{\fill}%  
  \begin{subfigure}{\textwidth}
    \includegraphics[height=0.22\textheight]{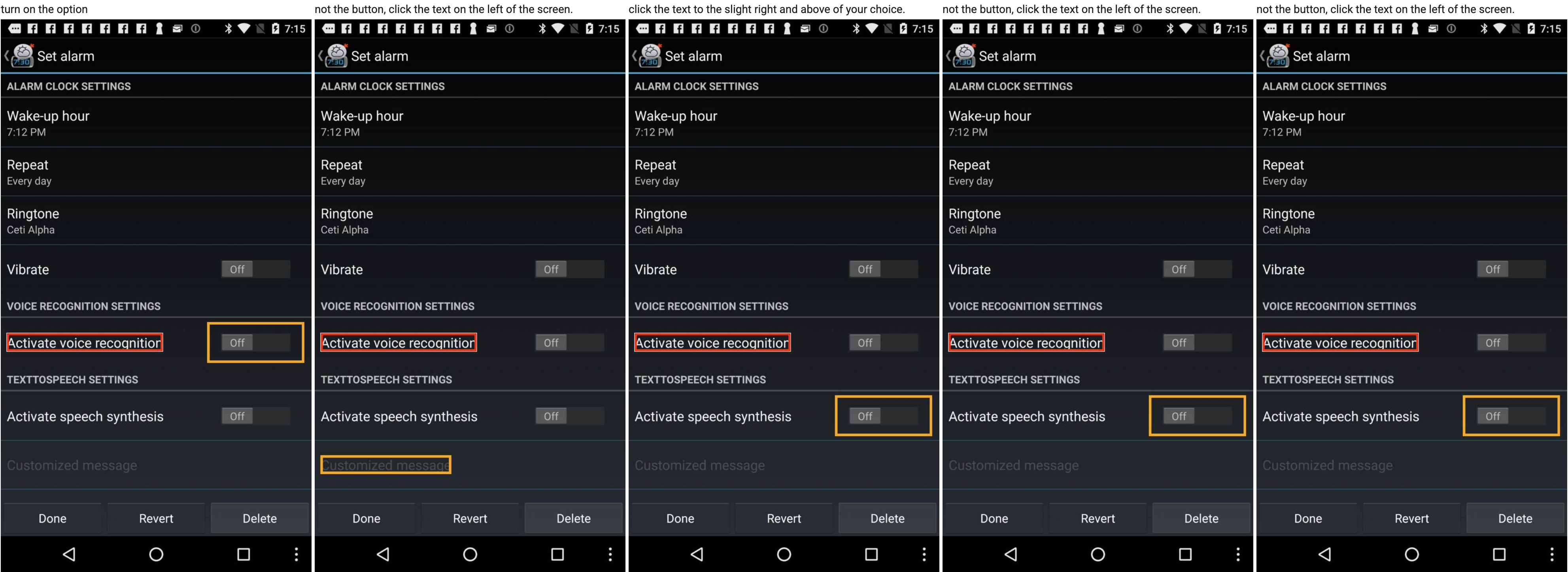}
    \label{fig:agent_view}
  \end{subfigure}%
  \hspace*{\fill}%
  
  \vspace*{2pt}%

  \hspace*{\fill}%  
  \begin{subfigure}{\textwidth}
    \includegraphics[height=0.22\textheight]{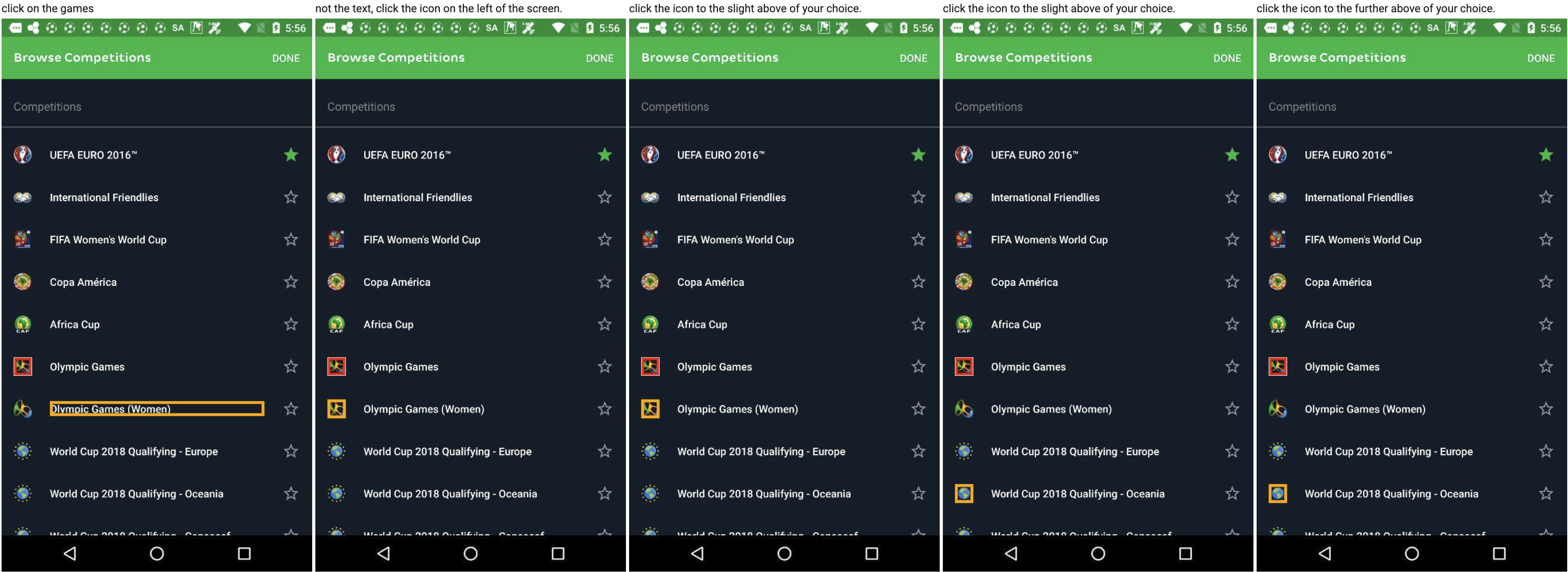}
    \label{fig:agent_view}
  \end{subfigure}%
  \hspace*{\fill}%
  
%  \captionsetup{skip=4pt, format=nocap}% % required to hide the figure (main) lable!
  \caption{\small{\tb{Failed} examples by the \emph{Imitation} agent following the instructions generated by the \tb{Heuristic} user.}}~\label{fig:failed_heu}
\end{figure}

\begin{figure}[ht]
 \centering
 \vspace*{5pt}%
 \hspace*{\fill}% 
  \begin{subfigure}{\textwidth}
    \includegraphics[height=0.22\textheight]{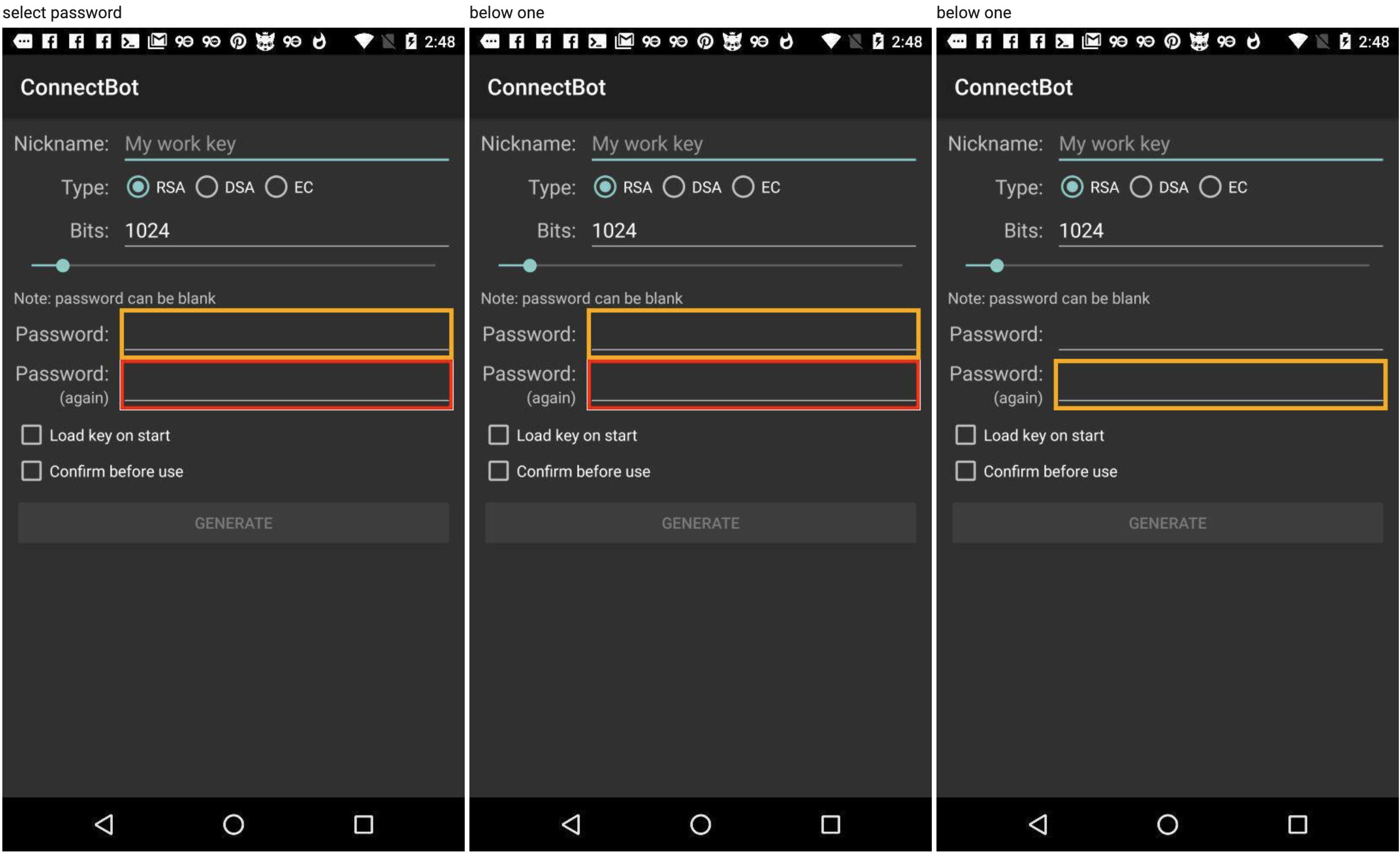}
    \label{fig:user_view}
  \end{subfigure}%
  \hspace*{\fill}%          % empty line absolutely necessary!

  \vspace*{2pt}%  

  \hspace*{\fill}%  
   \begin{subfigure}{\textwidth}
    \includegraphics[height=0.22\textheight]{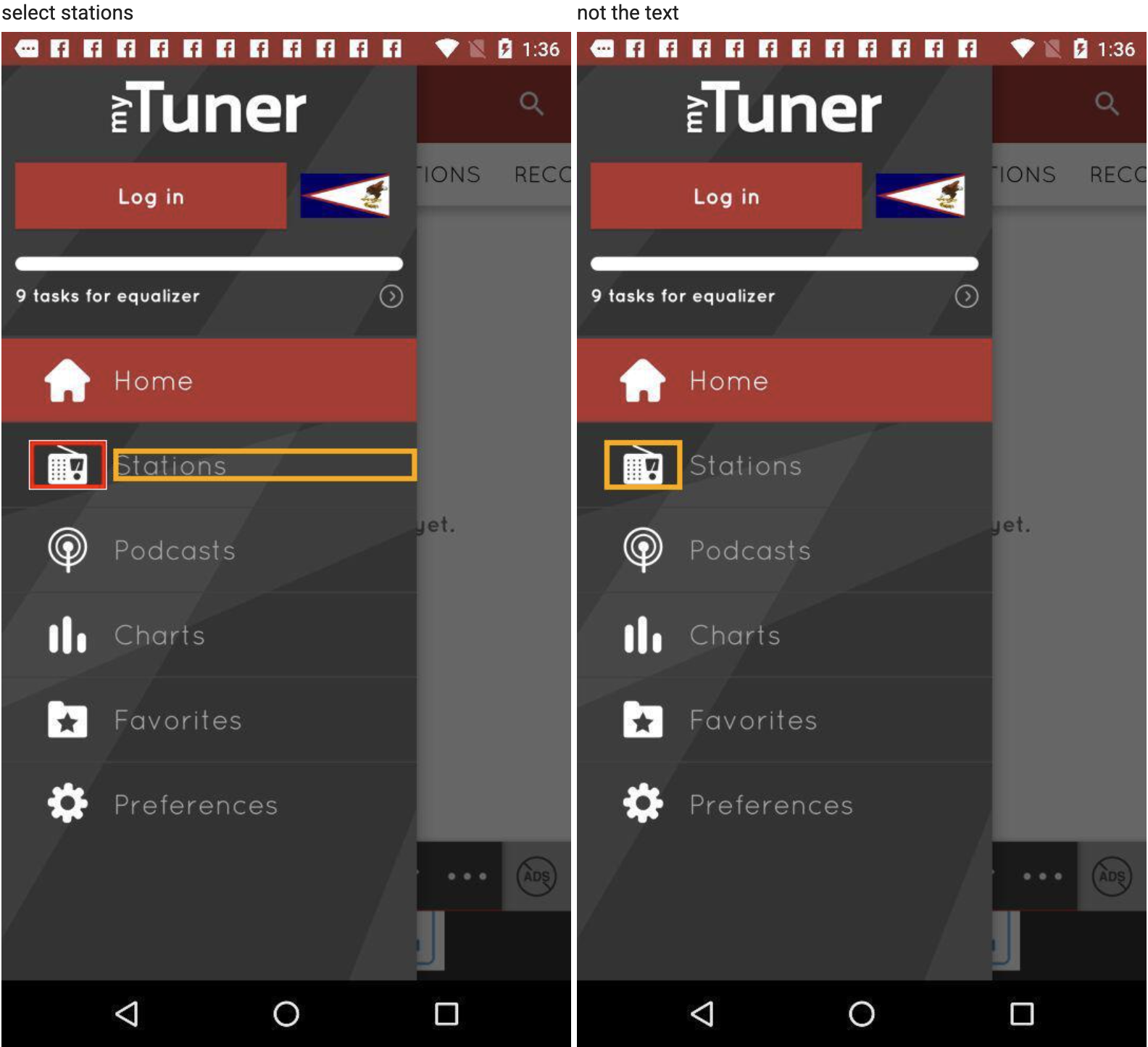}
    \label{fig:agent_view}
  \end{subfigure}%
  \hspace*{\fill}%          % empty line absolutely necessary!

  \vspace*{2pt}%

  \hspace*{\fill}%  
  \begin{subfigure}{\textwidth}
    \includegraphics[height=0.22\textheight]{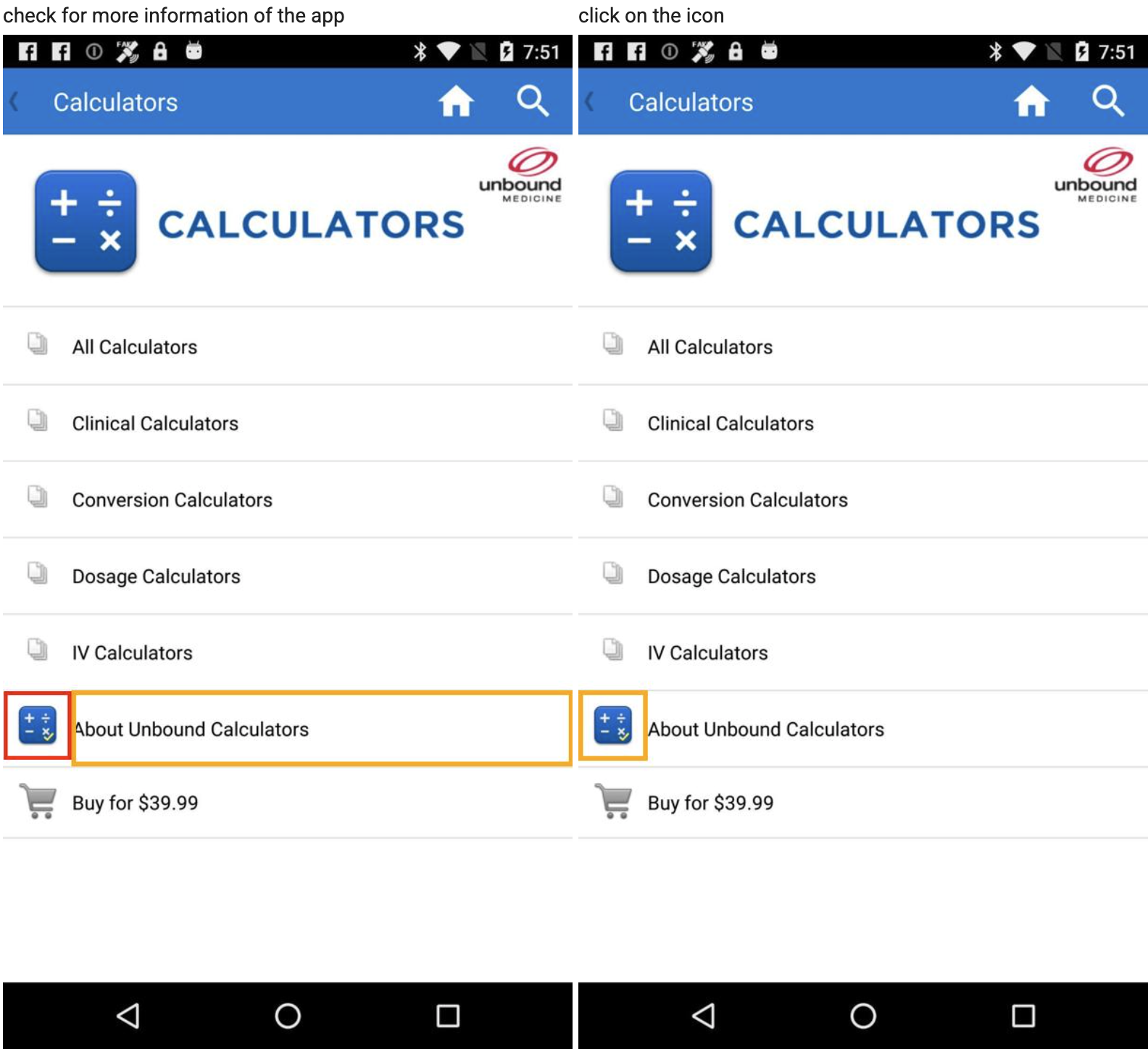}
    \label{fig:agent_view}
  \end{subfigure}%
  \hspace*{\fill}%
  
  \vspace*{2pt}%

  \hspace*{\fill}%  
  \begin{subfigure}{\textwidth}
    \includegraphics[height=0.22\textheight]{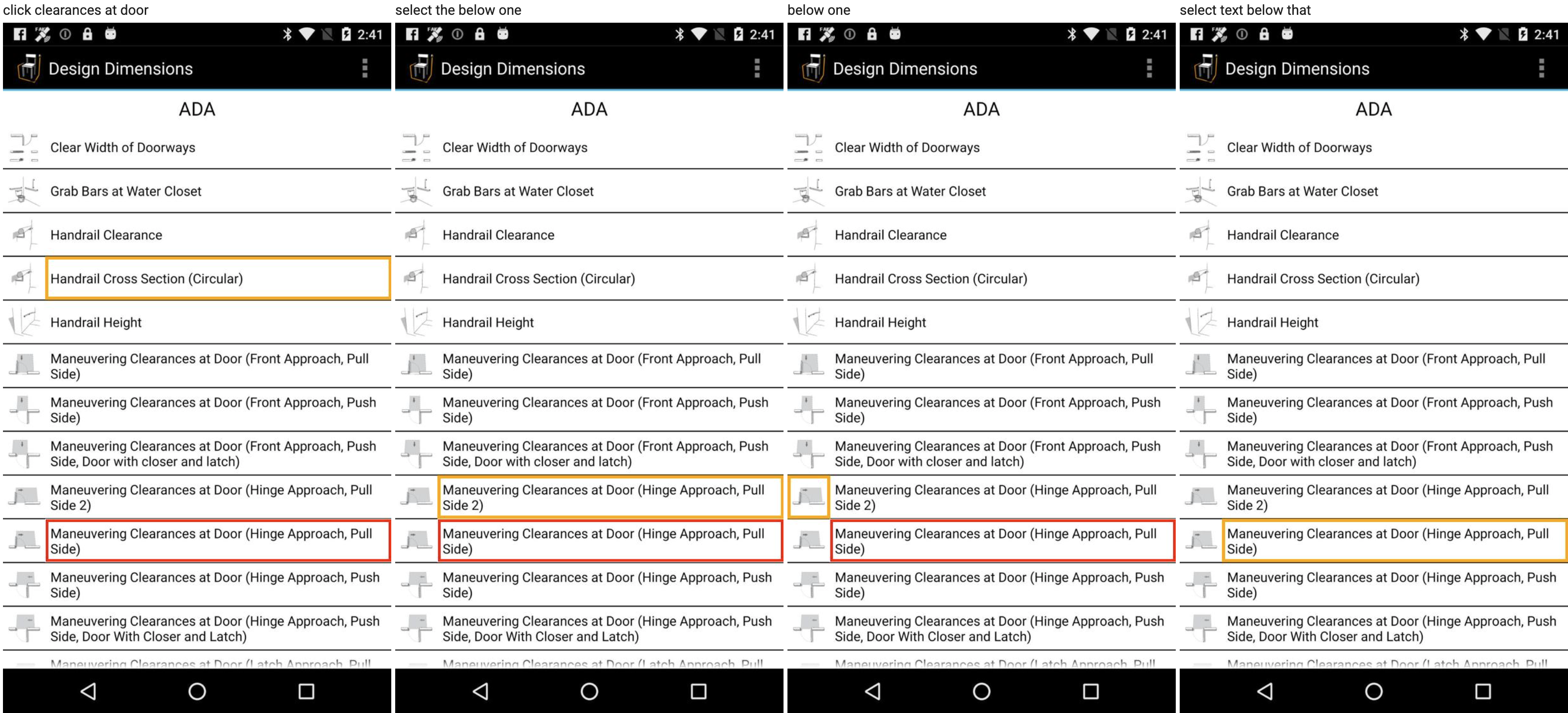}
    \label{fig:agent_view}
  \end{subfigure}%
  \hspace*{\fill}%
  
%  \captionsetup{skip=4pt, format=nocap}% % required to hide the figure (main) lable!
  \caption{\small{\tb{Completed} examples by the \emph{Imitation} agent following the instructions generated by the \tb{Neural} user.}}~\label{fig:completed_neural}
\end{figure}

\begin{figure}[ht]
 \centering
 \vspace*{5pt}%
 \hspace*{\fill}% 
  \begin{subfigure}{\textwidth}
    \includegraphics[height=0.22\textheight]{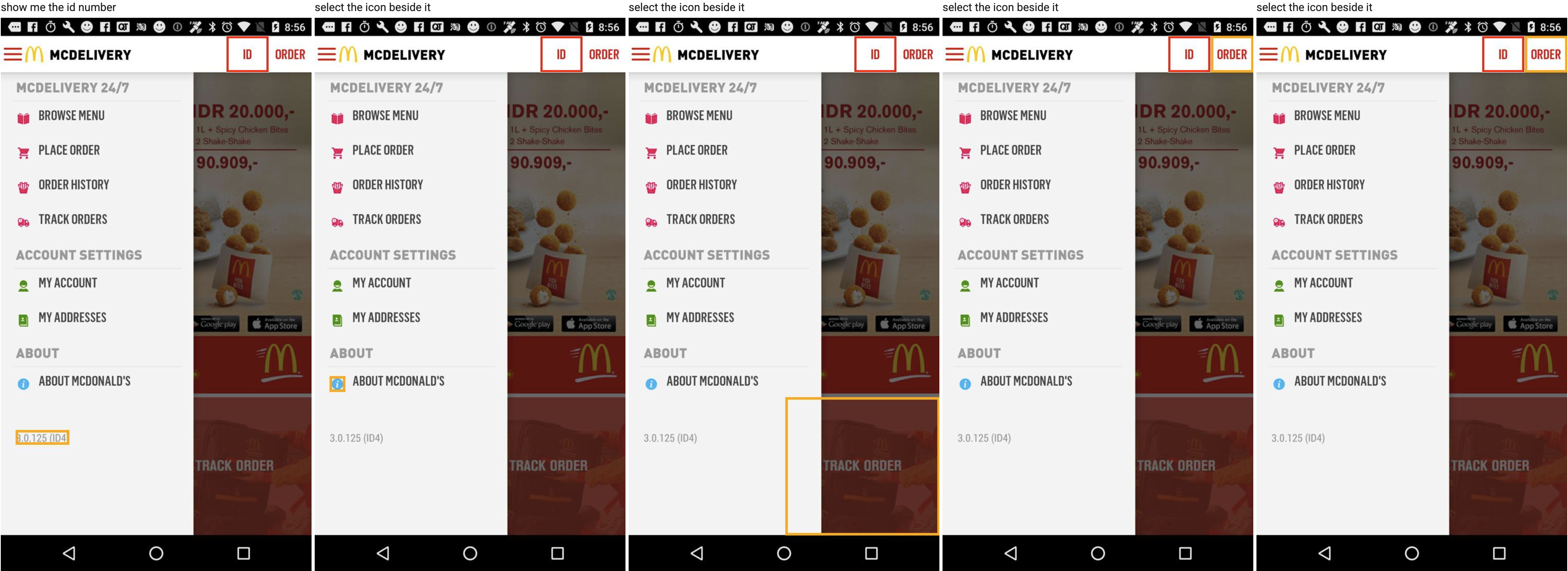}
    \label{fig:user_view}
  \end{subfigure}%
  \hspace*{\fill}%          % empty line absolutely necessary!

  \vspace*{2pt}%  

  \hspace*{\fill}%  
   \begin{subfigure}{\textwidth}
    \includegraphics[height=0.22\textheight]{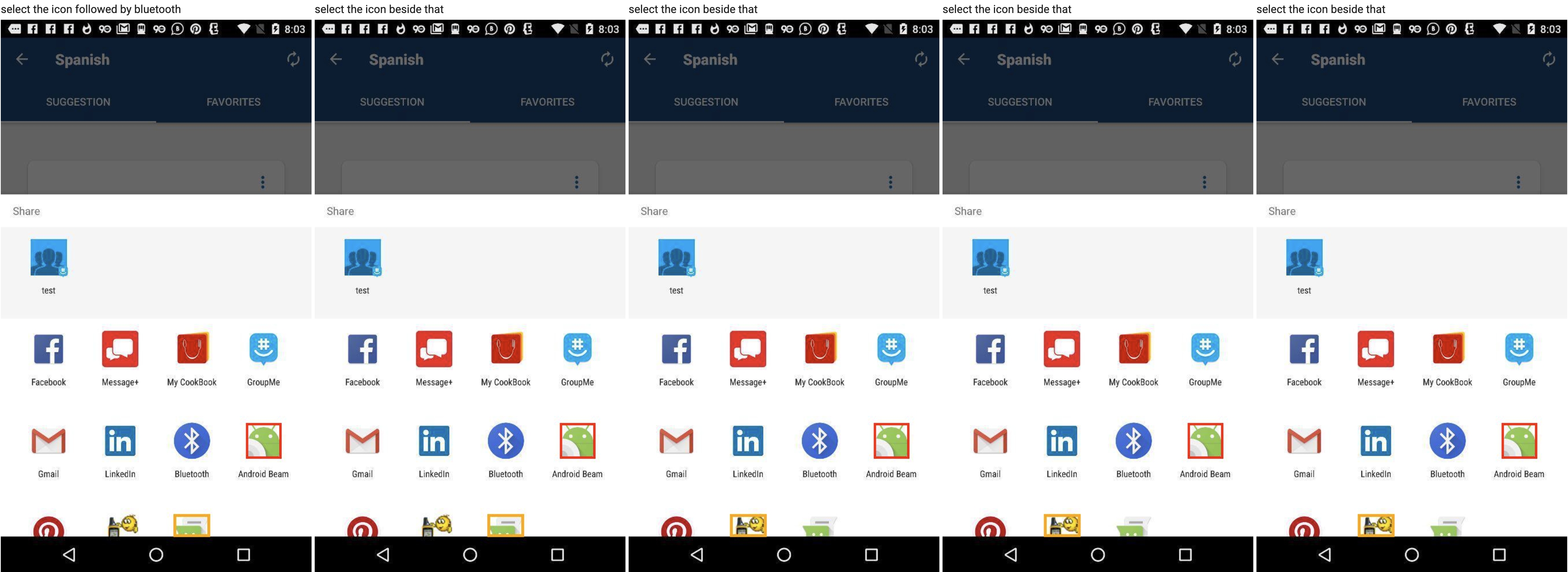}
    \label{fig:agent_view}
  \end{subfigure}%
  \hspace*{\fill}%          % empty line absolutely necessary!

  \vspace*{2pt}%

  \hspace*{\fill}%  
  \begin{subfigure}{\textwidth}
    \includegraphics[height=0.22\textheight]{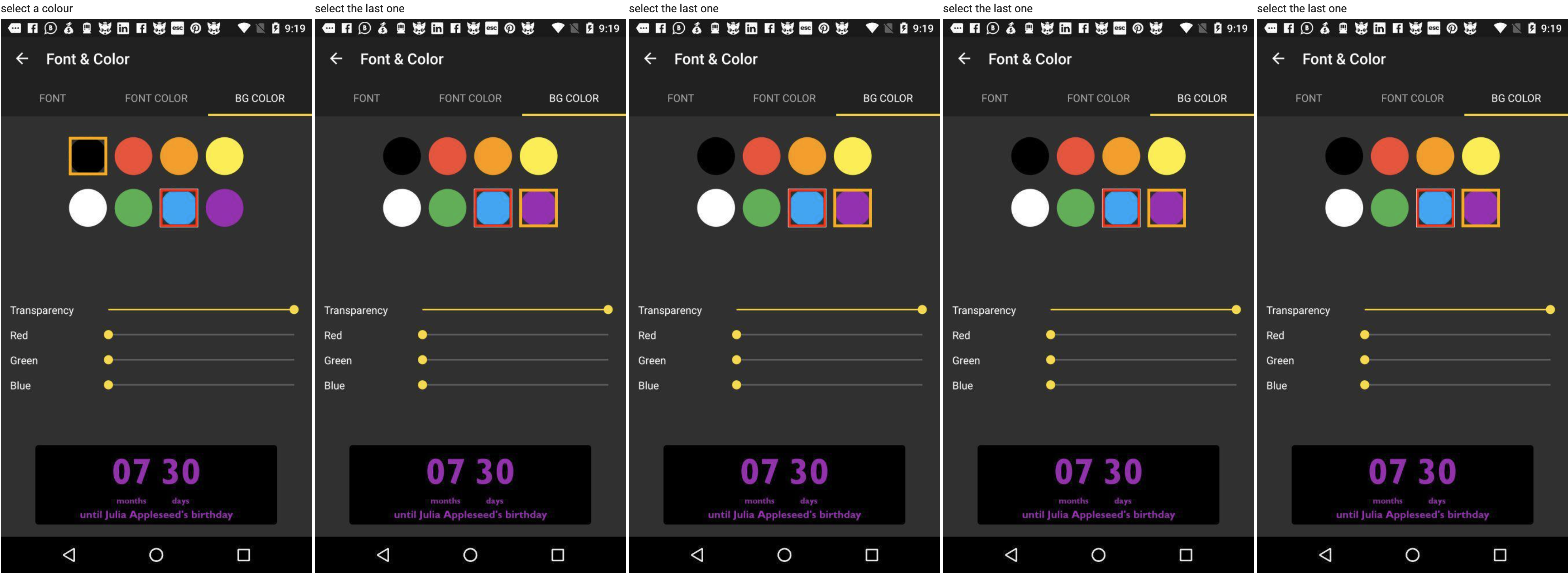}
    \label{fig:agent_view}
  \end{subfigure}%
  \hspace*{\fill}%
  
  \vspace*{2pt}%

  \hspace*{\fill}%  
  \begin{subfigure}{\textwidth}
    \includegraphics[height=0.22\textheight]{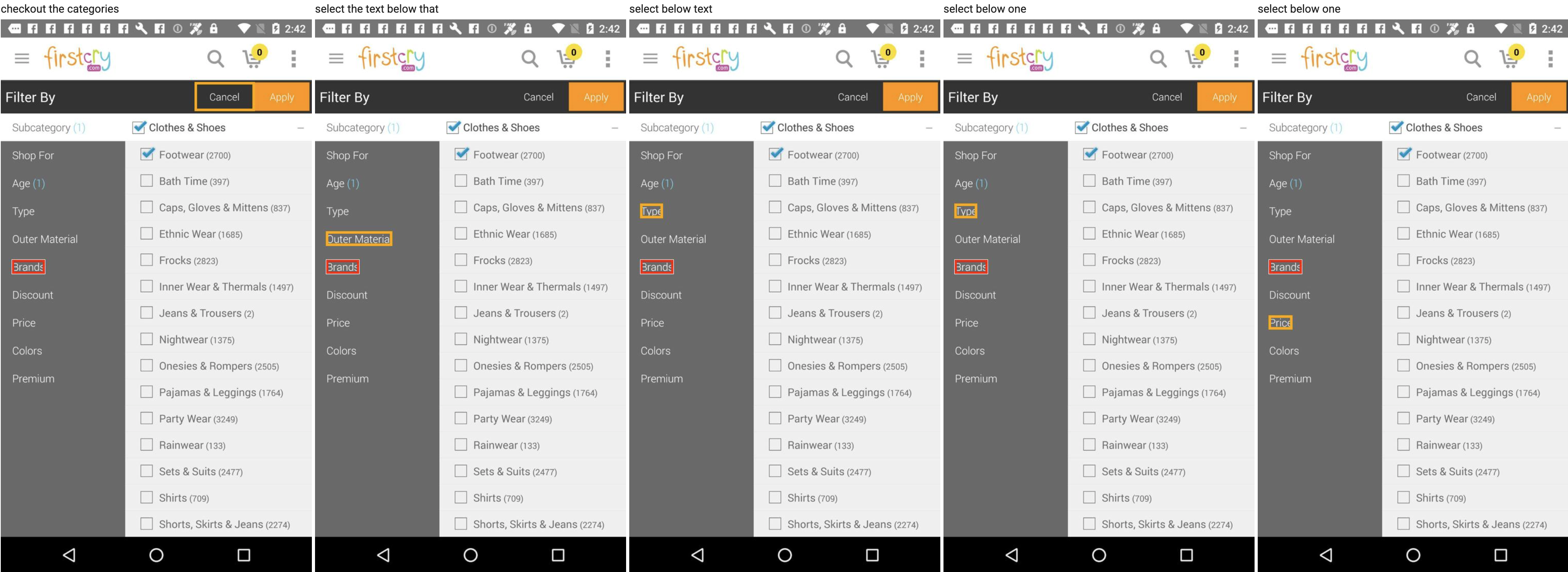}
    \label{fig:agent_view}
  \end{subfigure}%
  \hspace*{\fill}%
  
%  \captionsetup{skip=4pt, format=nocap}% % required to hide the figure (main) lable!
  \caption{\small{\tb{Failed} examples by the \emph{Imitation} agent following the instructions generated by the \tb{Neural} user.}}~\label{fig:failed_neural}
\end{figure}

Here, we demonstrate predictions from the \emph{Imitation} model.
Fig.~\ref{fig:completed_heu} demonstrates successfully solved examples following the instructions generated by the \emph{Heuristic} user model,
while failed ones are in Fig.~\ref{fig:failed_heu}.
Similarly, Fig.~\ref{fig:completed_neural} demonstrates solved ones following the instructions generated by the \emph{Neural} user model,
and failed ones are in Fig.~\ref{fig:failed_neural}.

